\documentclass[lettersize,journal]{IEEEtran}
\usepackage{amsmath,amsfonts}
\usepackage{algorithmic}
\usepackage{algorithm}
\usepackage{array}
\usepackage[caption=false,font=normalsize,labelfont=sf,textfont=sf]{subfig}
\usepackage{textcomp}
\usepackage{stfloats}
\usepackage{url}
\usepackage{verbatim}
\usepackage{graphicx}
\usepackage{cite}

\usepackage{booktabs}
\usepackage{color}
\usepackage{colortbl}
\usepackage{xcolor}
\usepackage{amssymb}

\graphicspath{{Figures-PDF/}}
\makeatletter
\def\endthebibliography{%
  \def\@noitemerr{\@latex@warning{Empty `thebibliography' environment}}%
  \endlist
}
\makeatother

\begin{document}

\title{Nighttime Thermal Infrared Image Colorization with Feedback-based Object Appearance Learning}

\author{Fu-Ya~Luo,
        Shu-Lin~Liu,
        Yi-Jun~Cao,
        Kai-Fu~Yang,
        Chang-Yong~Xie,
        Yong~Liu,
        and~Yong-Jie~Li,~\IEEEmembership{Senior~Member,~IEEE}
        % <-this % stops a space
\thanks{Fu-Ya Luo, Yi-Jun Cao, Kai-Fu Yang and Yong-Jie Li 
are with the MOE Key Laboratory for Neuroinformation, the School of Life Science and Tecnology, 
University of Electronic Science and Technology of China, Chengdu 610054, 
China. E-mail: luofuya1993@gmail.com, yijuncaoo@gmail.com, yangkf@uestc.edu.cn, 
liyj@uestc.edu.cn.}
\thanks{Shu-Lin Liu, Chang-Yong Xie and Yong Liu are with the Naval medical center of 
PLA, Shanghai, 200433, China. E-mail: shulinliu@smmu.edu.cn, xiechangyong@smmu.edu.cn, 
mike\_ly@163.com. (\textit{Corresponding author: Shu-Lin Liu, Yong Liu.})}
}

\markboth{Journal of \LaTeX\ Class Files,~Vol.~14, No.~8, August~2021}%
{Shell \MakeLowercase{\textit{et al.}}: A Sample Article Using IEEEtran.cls for IEEE Journals}


\maketitle

\begin{abstract}
  Stable imaging in adverse environments (e.g., total darkness) makes thermal infrared (TIR) 
  cameras a prevalent option for night scene perception. However, the low contrast 
  and lack of chromaticity of TIR images are detrimental to human interpretation and 
  subsequent deployment of RGB-based vision algorithms. Therefore, it makes sense to 
  colorize the nighttime TIR images by translating them into the corresponding daytime 
  color images (NTIR2DC). Despite the impressive progress made in the NTIR2DC task, how 
  to improve the translation performance of small object classes is under-explored. To 
  address this problem, we propose a generative adversarial network incorporating 
  feedback-based object appearance learning (FoalGAN). Specifically, an occlusion-aware 
  mixup module and corresponding appearance consistency loss are proposed to reduce the 
  context dependence of object translation. As a representative example of small objects 
  in nighttime street scenes, we illustrate how to enhance the realism of traffic light 
  by designing a traffic light appearance loss. To further improve the appearance 
  learning of small objects, we devise a dual feedback learning strategy to selectively 
  adjust the learning frequency of different samples. In addition, we provide 
  pixel-level annotation for a subset of the Brno dataset, which can facilitate the 
  research of NTIR image understanding under multiple weather conditions. Extensive 
  experiments illustrate that the proposed FoalGAN is not only effective for 
  appearance learning of small objects, but also outperforms other image translation 
  methods in terms of semantic preservation and edge consistency for the NTIR2DC task.
\end{abstract}

\begin{IEEEkeywords}
  Thermal infrared image colorization, image-to-image translation, generative adversarial 
  networks, feedback-based learning, nighttime scene perception.
\end{IEEEkeywords}

\section{Introduction}
\IEEEPARstart{N}{ighttime} scene perception is a fundamental yet challenging 
computer vision tasks. Despite its superiority in texture capture, visible spectrum-based 
perception systems fail in total darkness conditions and do not enable accurate imaging of 
dazzling areas. On the contrary, the insensitivity to lighting conditions and strong 
penetration in hazy environments 
make thermal infrared-based imaging one of the promising solutions for nighttime 
scene perception \cite{2018-TCSVT-Li,2016-TCSVT-Li}. However, the low contrast and monochromatic nature of thermal infrared 
(TIR) images hinder human interpretation \cite{2009-EOJ-W} and subsequent deployment of RGB-based models. 
Therefore, the translation from nighttime TIR images into corresponding daytime color (DC) 
images (abbreviated as NTIR2DC) is of great significance, which not only benefits the rapid understanding of 
nighttime scenes, but also reduces the annotation cost of NTIR image analysis tasks through 
existing visible-based datasets and algorithms.

Due to the difficulty of acquiring a large number of pixel-level aligned cross-domain image 
pairs, more efforts \cite{2022-TITS-Luo,2021-ICIG-Luo} have been devoted to the 
implementation of NTIR2DC tasks using unsupervised 
image-to-image (I2I) translation methods. Unsupervised I2I 
translation aims to learn mappings between different image domains under the condition of 
unpaired samples \cite{2017-CVPR-Isola}. In recent years, driven by the promising advances in 
image generation, generative adversarial networks (GANs) \cite{2014-NIPS-Goodfellow} have 
attracted extensive attention in the field of unsupervised I2I translation. For example, 
Zhu \textit{et al.} \cite{2017-CVPR-Zhu} proposed CycleGAN to encourage cycle-consistency 
between the reconstructed image after inverse translation and the original image. To enhance 
the diversity of translation results, \cite{2018-ECCV-Huang} and 
\cite{2020-IJCV-Lee} learned separated image content representations and style representations. 
ToDayGAN \cite{2019-ICRA-Anoosheh} utilized multiple discriminators to improve night-to-day 
translation performance. PearlGAN \cite{2022-TITS-Luo} introduced a top-down guided attention 
module and corresponding losses to reduce semantic encoding entanglement in the NTIR2DC task.

\begin{figure}[!t]
\centering
\includegraphics[width=3.45in]{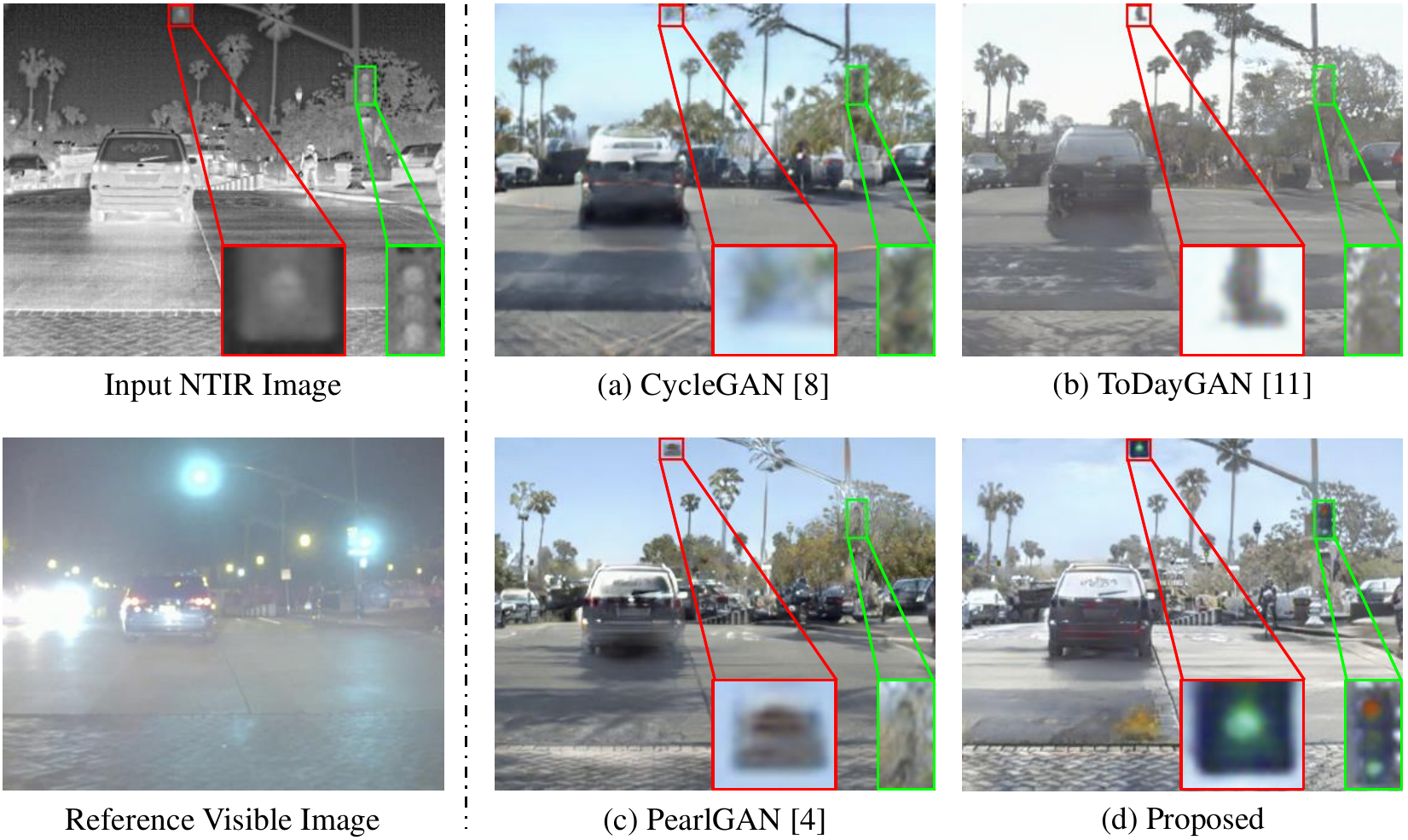}
\caption{Visual comparison of colorization results. The areas in the red and 
green boxes deserve attention.}
\label{fig_intro}
\end{figure}

Despite the impressive results achieved by previous approaches, the problem of how to mitigate 
the context dependence of object translation is under addressed. Moreover, it is 
under-explored how to improve the colorization performance of small object categories (SOC) 
without semantic annotation. As shown in Fig. \ref{fig_intro}, previous methods consistently 
failed to generate plausible traffic lights. To handle these limitations mentioned above, a 
GAN network incorporating Feedback-based Object Appearance Learning, referred to as FoalGAN, 
is proposed in this paper.

In the NITR2DC task, we observed that there may be large differences in the translation 
results of same objects under different contextual conditions, which is defined by us as the 
context dependence of object translation. Inspired by Mixup \cite{2018-ICLR-Zhang} in 
the enhancement of model generalization, we devise an occlusion-aware mixup (OAMix) module to 
fuse fake objects with real images, constraining the consistency of object translation through 
an appearance consistency loss to reduce context dependence. Unlike previous approaches, the 
proposed OAMix utilizes pseudo-labels to avoid the occlusion of cross-domain objects, and an 
adaptive luminance adjustment strategy is designed to enhance the naturalness of the 
synthesized NTIR images. 

For small object translation, lack of supervision and low 
percentage contributed to the loss of objective function make feature learning of SOC difficult, and the random sample 
learning strategy of GAN networks worsens this issue. Compared with artificial neural networks, humans have 
powerful learning capabilities that allow them to master a wide range of skills, and feedback 
during learning process can help learners improve the systematicity of their strategies 
to better acquire and apply 
knowledge \cite{2005-LI-Vollmeyer,2014-BBR-Luft}. Motivated by this feedback-based learning 
scheme, we propose a dual feedback learning strategy to improve the translation performance 
of SOC. Specifically, the appearance consistency loss of the current iteration modulates 
the input sample selection for the next iteration through a feedback connection, which 
adaptively adjusts the learning frequency of small object samples in both domains.

In addition, to illustrate how to facilitate the translation of small objects with the 
proposed framework, we select the traffic light as a representative example in nighttime 
street scenes to show how to enhance the colorization performance of such object by 
designing an elaborate traffic light appearance loss. Furthermore, regarding the 
evaluation of NTIR2DC methods, previous methods have been studied almost 
exclusively on NTIR datasets under cloudy conditions, what if the images are collected under 
rainy conditions? And what if the training set images contain multiple types of weather? 
To explore these concerns, we conduct further experiments on the Brno \cite{2020-ICRA-Ligocki} 
dataset containing NTIR images of multiple weather conditions. Further, we sample a subset for 
pixel-level semantic annotation. Exhaustive experiments show that the proposed method not 
only improves the colorization performance of SOC, but also significantly 
outperforms existing translation methods in terms of semantic and edge preservation. The 
main contributions of this study are:
\begin{itemize}
  \item An OAMix module and the corresponding appearance consistency loss are 
  designed to reduce the context dependence of object translation and enhance the 
  generalization capacity of model.
  \item A dual feedback learning strategy is proposed to adaptively adjust the learning frequency 
  of small object samples, which is beneficial to improve the translation performance of small 
  objects.
  \item A traffic light appearance loss considering both luminance and color is proposed to 
  enhance the overall realism of the traffic light, which may provide inspiration to enhance 
  specific small objects under specific conditions for specific tasks.
  \item We annotate a subset of the Brno dataset with semantic masks, which may catalyze the 
  researches of NTIR image understanding for multiple weather conditions.
  \item The experimental results of the NTIR2DC task demonstrate the superiority of the 
  proposed FoalGAN\footnote{The source code will be available 
  at \url{https://github.com/FuyaLuo/FoalGAN/}.} 
  in terms of small-object colorization and semantic consistency of translation.
\end{itemize}

The rest of this paper is organized as follows. Section II summarizes related work about TIR image colorization and I2I 
translation. Section III introduces the architecture of the proposed FoalGAN. Section IV presents 
the experiments on the FLIR and Brno datasets. Section V draws the conclusions.

\section{Related Work}
In this section, we briefly review previous work on TIR image colorization, unpaired 
I2I translation, neural networks with feedback connections, and mask-based mixup strategy.
\subsection{TIR Image Colorization}
TIR image colorization aims to map a grayscale TIR image to a three-channel RGB image without 
changing semantics. Based on the demand for paired RGB images, TIR image colorization 
methods can be divided into two categories: supervised and unsupervised. The supervised 
approach forces the model to maximize the similarity of its output to the aligned RGB image. 
For example, Berg \textit{et al.} \cite{2018-Berg-CVPRW} realized TIR image colorization by 
introducing separate luminance and chrominance loss. In order to increase the naturalness of 
the results, more efforts \cite{2020-Bhat-ICCES,2020-Kuang-IPT,2021-ISEE-Le} have focused on 
combining pixel-level content loss with adversarial loss to optimize the mapping from TIR to 
RGB images. SCGAN \cite{2020-TCSVT-Zhao} jointly predicted the colorization and saliency map 
to reduce semantic confusion and color bleeding.
Compared with supervised methods, unsupervised methods do not require finely 
aligned cross-domain samples, which greatly reduces data collection costs and extends their 
applicability. Most unsupervised colorization methods utilize the GAN model to implement 
the image translation from TIR to visible spectrum. For example, 
Nyberg \textit{et al.} \cite{2018-Nyberg-ECCV} exploited the CycleGAN \cite{2017-CVPR-Zhu} 
model to achieve unsupervised TIR image colorization. PearlGAN \cite{2022-TITS-Luo} was 
proposed to reduce semantic encoding entanglement and edge distortion in the NTIR2DC task. 
DlamGAN \cite{2021-ICIG-Luo} introduced a dynamic label mining module and a segmentation 
loss to encourage the semantic consistency of NTIR colorization process. To enhance the 
semantic preservation of small sample categories, MornGAN \cite{2022-Arxiv-Luo} introduced 
a memory-guided collaborative attention strategy. Despite their encouraging progress, few 
efforts have been made to improve the colorization performance of SOC.

\subsection{Unpaired I2I Translation}
The purpose of unpaired I2I translation is to learn a transformation function through 
unpaired samples to achieve cross-domain mapping of images. Zhu \textit{et al.} \cite{2017-CVPR-Zhu} 
made the earliest effort to get rid of the requirement of aligned image pairs by using cycle consistency loss, 
leading to the recent surge of interest in methods for unpaired I2I translation. On the 
basis of CycleGAN, UNIT \cite{2017-NIPS-Liu} replaced the domain-specific latent space with 
a shared latent space. Recent works further boosted the generation performance of night-to-day 
image translation by introducing multiple discriminators \cite{2019-ICRA-Anoosheh}, additional 
decoders \cite{2020-ECCV-Zheng}, or auxiliary detection 
branches \cite{2020-CVPR-Bhattacharjee,2021-TCSVT-Fu}. In order to reduce content distortion 
during translation, semantic consistency loss is more often introduced into GAN 
models \cite{2018-ECCV-Huang-Auggan,2019-WACV-Cherian}. Considering the demand for 
incremental learning in practical applications, Tan \textit{et al.} \cite{2020-TCSVT-Tan} 
proposed an IncrementalGAN that can incrementally learn new domains without requiring the 
training data from the previously learned domains. Although these approaches have obtained 
impressive results, how to improve the translation performance of SOC under 
unlabeled conditions remains under-explored. In addition, the issue of how to reduce the 
context dependence of object translation has been little studied.

\subsection{Neural Networks with Feedback Connections}
Feedback is defined as an event that reroutes the (partial or full) output of a system back into 
the input and as part of an iterative cause-and-effect process \cite{1999-IP-Ford}. 
Feedback connections are beneficial for humans to achieve flexible pattern recognition in 
complex and rapidly changing 
environments \cite{1998-Nature-Hupe,2007-Neuron-Gilbert,2012-JCogNeu-Wyatte}. Inspired by 
this, many researchers \cite{2014-NeurIPS-Stollenga,2017-CVPR-Zamir} have introduced 
feedback connections in the design of deep networks. For example, 
DasNet \cite{2014-NeurIPS-Stollenga} utilized a feedback structure to dynamically adjust 
the sensitivity of each convolutional filter during classification. To achieve iterative 
inference, some studies \cite{2017-CVPR-Zamir,2019-CVPR-Li} employed a convolutional 
recurrent neural network model to pass the output of each iteration through a hidden 
state to the next iteration. Song \textit{et al.} \cite{2022-TCSVT-Song} incorporated a feedback 
mechanism into the model to reuse the intermediate results of each stream to achieve 
single image dehazing. Unlike previous approaches that focus on feature integration, 
the proposed feedback connection in this work collaboratively adjusts the selection 
of input samples 
for both domains based on the learning state of model, which does not increase the 
computational and time costs of model inference.

\subsection{Mask-based Mixup Strategy}
Mixup is a data augmentation strategy that combines sample pairs and their labels in 
a linear fashion \cite{2018-ICLR-Zhang}. Driven by its success in improving model 
generalization, the mask-based mixup strategy has been widely adopted for various 
computer vision tasks \cite{2019-ICCV-Yun,2021-WACV-Olsson,2022-TCSVT-Zhou}. For 
example, CutMix \cite{2019-ICCV-Yun} 
cut and pasted a rectangular box area from one image to another. 
In the unlabeled condition, ClassMix \cite{2021-WACV-Olsson} 
cut and pasted half of the classes of one image to another image based on the predicted 
masks.  
RICAP \cite{2019-TCSVT-Takahashi} stitched four images into one image after cropping out 
small pieces at a certain ratio. Considering the cross-domain context dependency, 
CAMix \cite{2022-TCSVT-Zhou} first used the spatial distribution of the source domain and 
the contextual relationship of the target domain to estimate the contextual mask, and then 
performed subsequent domain mixup based on the mask. Despite the 
remarkable progress of the mask-based mixup strategy, how to avoid the unrealistic 
problem (e.g., a bus appears in the abdomen of a pedestrian) caused by object occlusion 
in fused images under unlabeled conditions remains to be explored. In addition, how to 
reduce the luminance differences between patches of TIR image mixup is under-resolved.

\section{Proposed Method}
In this section, we first present the overview of the proposed FoalGAN. Subsequently, we briefly 
explain our baseline model. Then, the details of bias correction in the fake NTIR image 
branch are described. Next, we explicate the OAMix module. Afterward, 
the traffic light appearance loss enhancing the appearance realism of traffic 
lights is explained. Then, the dual feedback learning strategy is specified. Finally, 
we illustrate the total loss of FoalGAN.

\begin{figure*}[!t]
\centering
\includegraphics[width=1\textwidth]{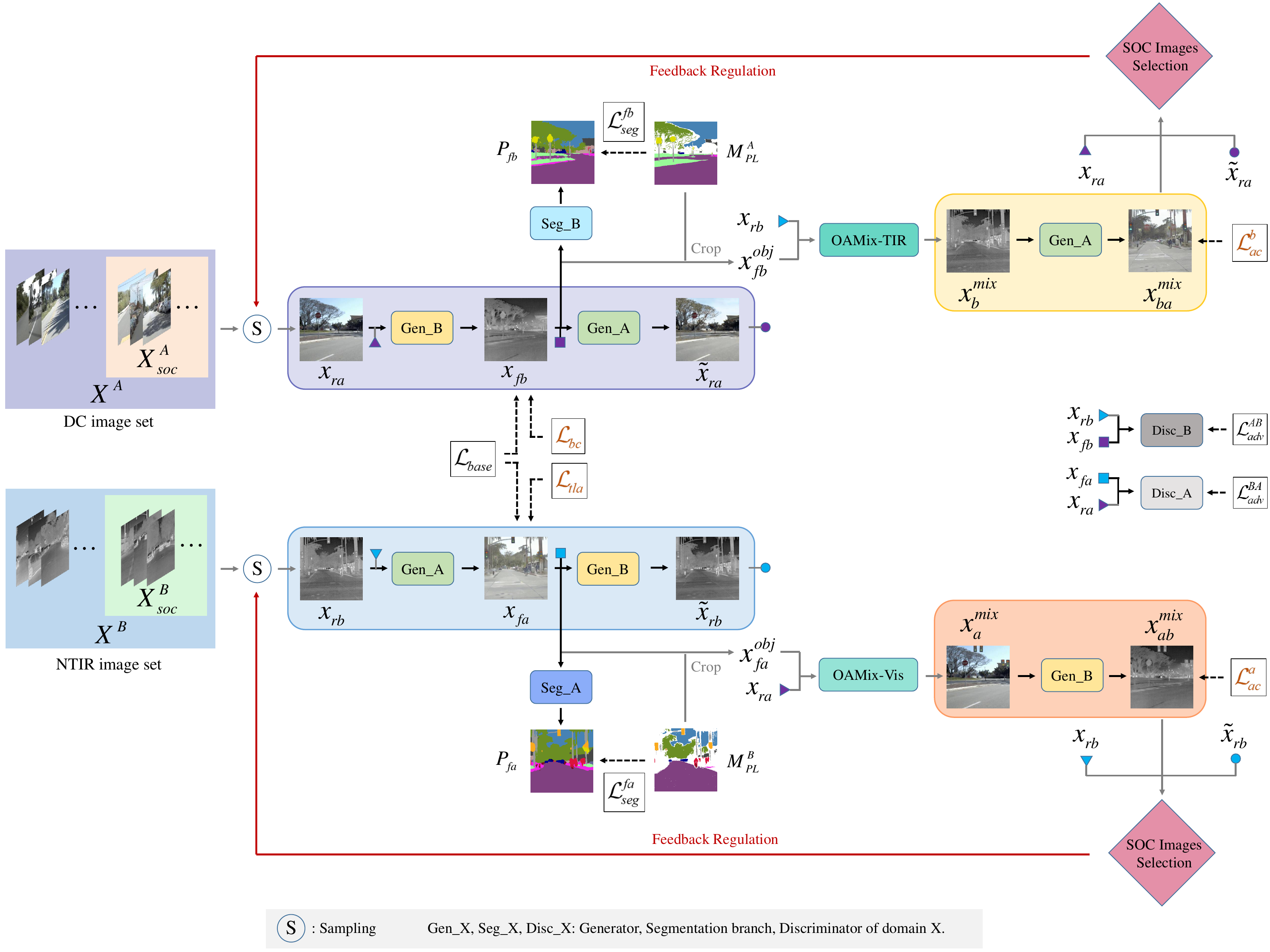}
\caption{The overall architecture of the proposed method. $x_{ra}$ and $x_{rb}$ are one image in 
the DC and NTIR domains, respectively, whose sampling is modulated by the feedback from the 
SOC image selection module of the previous iteration. The translation results of the input 
image pairs, denoted as $x_{fb}$ and $x_{fa}$, are first mixed with the real 
images (i.e., $x_{rb}$ and $x_{ra}$) through the OAMix module under the 
guidance of pseudo-labels (i.e., $M_{PL}^{A}$ and $M_{PL}^{B}$). The corresponding appearance 
consistency loss (i.e., $\mathcal{L}_{ac}^{a}$ and $\mathcal{L}_{ac}^{b}$) then encourages 
the robustness of the SOC translation against background variations. 
And the samples containing SOC in the DC and NTIR domains are denoted as $X_{soc}^{A}$ 
and $X_{soc}^{B}$, respectively.}
\label{fig_model}
\end{figure*} 

\subsection{Model's Overview and Problem Formulation}
The overall framework is shown in Fig. \ref{fig_model}. We first choose MornGAN \cite{2022-Arxiv-Luo} as 
baseline model to encourage semantic consistency of translation, and its total loss is 
abbreviated as $\mathcal{L}_{base}$. Then, we introduce a bias correction loss $\mathcal{L}_{bc}$ 
to reduce the bias of the fake NTIR image branch. Next, we devise an OAMix module and an 
appearance consistency loss $\mathcal{L}_{ac}$ to moderate the context dependence 
of SOC translation. Subsequently, a traffic light appearance 
loss $\mathcal{L}_{tla}$ is designed to enhance the realism of traffic lights in 
colorization results. Finally, we propose a dual feedback learning strategy to rationally 
allocate the learning frequency of samples containing SOC, which aims to improve the 
translation performance of SOC while reducing the degradation of learning for other categories.

In the rest of the paper, domain A and domain B denote the DC image set and NTIR image set, 
respectively, abbreviated as $X^{A}$ and $X^{B}$. Taking the translation from domain A to 
domain B as an example, we denote the input 
image pair of domain A and B as $\left \{ x_{ra}, x_{rb} \right \} $, the 
generator $G_{AB}$ contains an encoder of domain A and a decoder of domain B, and the 
discriminator $D_B$ aims to distinguish the real image $x_{rb}$ from the translated 
image $x_{fb}$ (i.e., $G_{AB} \left (x_{ra} \right ) $). Similarly, the inverse mapping includes 
generator $G_{BA}$ and discriminator $D_A$.

\subsection{Baseline Model}
MornGAN \cite{2022-Arxiv-Luo} 
is proposed to improve the colorization performance of small sample categories in the NTIR2DC 
task. Specifically, two semantic segmentation networks (i.e., $S_{A}$ and $S_{B}$) and 
segmentation 
losses (i.e., $\mathcal{L}_{seg}^{fa}$ and $\mathcal{L}_{seg}^{fb}$) are introduced to 
encourage the semantic consistency of translation. Due to the lack of semantic 
annotation, two existing segmentation models (i.e., Detectron2 \cite{2019-detectron2-Wu} 
and HMSANet \cite{2020-Arxiv-Tao}) are first used to predict the pseudo-labels ($M_{PL}^{A}$) 
of DC images. The pseudo-labels ($M_{PL}^{B}$) of NTIR images are then obtained by an 
online semantic distillation module, which is subsequently stored in a memory unit. 
When the stored quantity meets the condition, the memory-guided sample selection strategy 
is triggered, and then similar NTIR images are recalled for the input DC images for 
collaborative learning. Finally, adaptive collaborative attention loss is used to encourage the 
inter-domain similarity of features of small sample categories. In addition, the conditional 
gradient repair loss is introduced to reduce edge degradation during translation, and a 
scale robustness loss is used to improve the insensitivity of the model to the object scale.

\subsection{Bias Correction in the Fake NTIR Image Branch}
\begin{figure}[!t]
\centering
\includegraphics[width=3.45in]{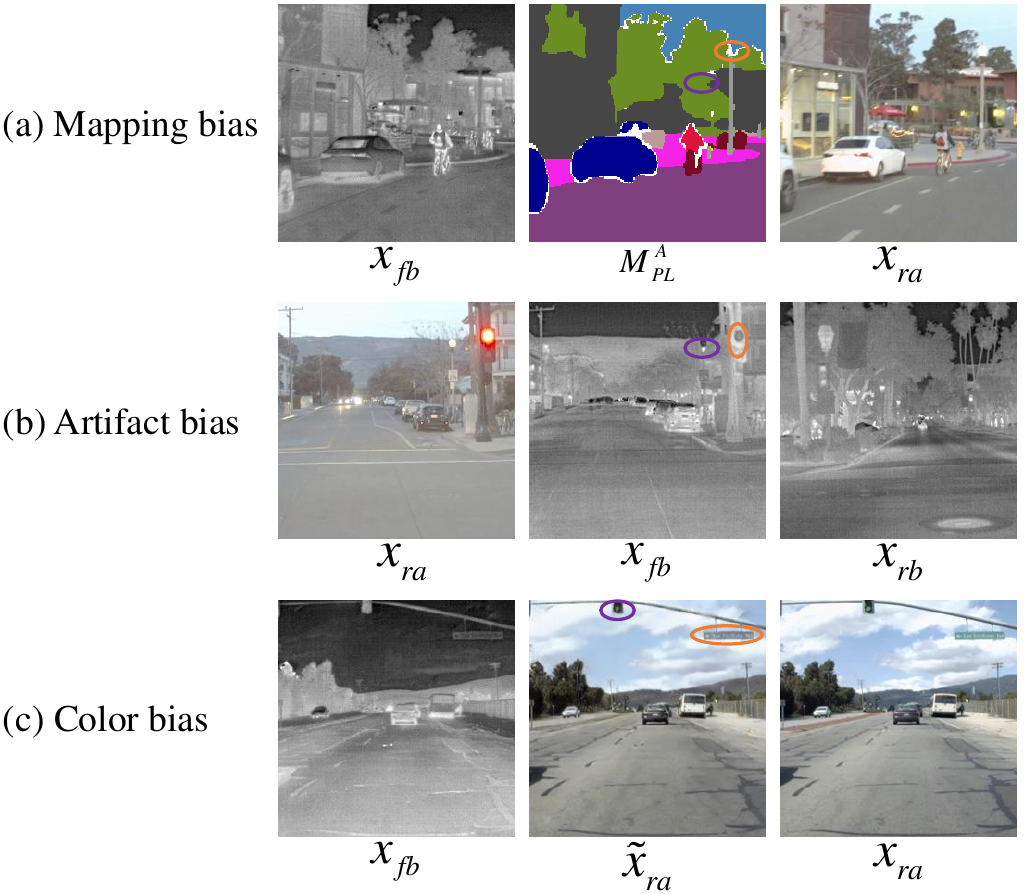}
\caption{Example images of biases in the fake NTIR image branch. The orange and purple boxes are 
two examples of the corresponding bias. $x_{ra}$, $x_{rb}$, $x_{fb}$, $\tilde{x}_{ra}$, 
and $M_{PL}^{A}$ denote the real DC image, the real NTIR image, the fake NTIR image, 
the reconstructed DC image, and the pseudo-labels of domain A, respectively.}
\label{fig_bc}
\end{figure}

Although MornGAN can effectively improve the colorization performance of small sample 
categories, how to reduce the various types of biases in the fake NTIR image branch during training is 
under-resolved. As shown in 
Fig. \ref{fig_bc}, the biases in the fake NTIR branch are divided into three main types: mapping 
bias of semantic segmentation, artifact bias of image translation and color bias of 
image reconstruction. Since the two branches in the model are highly coupled, the presence of 
these biases is detrimental to the appearance learning of SOC in both domains. 
In this section, we focus on how to correct for these biases.

\subsubsection{Mapping Correction for Semantic Segmentation}
Due to the lack of semantic annotation and the existance of some artifacts in the fake NTIR images, 
some mapping bias occurs in the semantic learning of fake NTIR images. Among them, 
there are two mapping biases that are highly relevant to the semantic prediction of 
SOC. The first bias is mapping the high luminance regions of the fake NTIR images to 
the vegetation category, and the second is mapping the streetlight regions to other categories, 
as shown in Fig. \ref{fig_bc}(a).

In order to reduce the first mapping bias, we set the semantic labels 
corresponding to regions with high luminance (i.e., greater than the luminance mean) 
in the vegetation category in the fake NTIR images to ``uncertain" (i.e., no loss calculation). 
For the second bias, we first add streetlight to the definition of the category set. Then, 
the masks of SOC for DC images are predicted by the HMSANet \cite{2020-Arxiv-Tao} trained 
on the Mapillary \cite{2017-ICCV-Neuhold} dataset, which is subsequently used to update the 
pseudo-labels $M_{PL}^{A}$.

\subsubsection{Artifact Correction for Image Translation}
Compared with supervised I2I translation, unpaired I2I translation usually produces local 
artifacts due to the lack of pixel-level labels. 
In general, the temperature of the luminous areas in street and traffic lights is usually 
higher than that of the neighboring regions under nighttime conditions. 
However, when no specific strategy is taken into account, bright areas in DC images belonging to streetlights and 
traffic lights become darker after translation, which deviates from the temperature 
distribution in the nighttime scenes, as shown in Fig. \ref{fig_bc}(b). Therefore, we propose an artifact bias correction 
loss $\mathcal{L}_{abc}$ to reduce such limitation. In order to correct the luminance 
bias of street lights, we propose a street light luminance adjustment loss $\mathcal{L}_{sla}$. 
We have observed that the temperature of luminous street light is 
usually higher than that of vegetation, e.g., the trees along the street. Motivated by this observation, the loss $\mathcal{L}_{sla}$ 
aims to encourage that the minimum value of the bright areas of streetlight in the fake NTIR 
image should be greater than the mean value of vegetation. 
Specifically, given the minimum value of the bright areas of the streetlight category in 
the fake NTIR image, denoted as $\tilde{t}_{fb}^{sl}$, and the mean value of the 
vegetation category, denoted as $\bar{t}_{fb}^{veg}$, the loss $\mathcal{L}_{sla}$ is defined as
\begin{equation}
  \label{L_sba}
  \mathcal{L}_{sla}=\max \left ( \bar{t}_{fb}^{veg}-\tilde{t}_{fb}^{sl}+  \theta_{tem},0 \right ),  
\end{equation}
where $\theta_{tem}$ is a threshold value to control the temperature difference between classes.

Due to the spatial hierarchy of the luminance distribution of traffic lights, a simple 
minimum constraint to reduce the artifact bias of traffic lights is sub-optimal. Therefore, 
inspired by the observation that the temperature of luminous region is higher 
compared with that of other regions, we propose another traffic light luminance adjustment 
loss $\mathcal{L}_{tla}$ to encourage the consistency of the luminance distribution during 
translation. Specifically, given a grayscale map of DC image 
as $I_{ra}^{g}\in \mathbb{R}^{H\times W}$, we first subtract the mean of traffic light 
region, then perform dot product using the binary mask of traffic light, and finally 
perform vectorization to obtain the result noted as $V_{ra}^{tl}\in \mathbb{R}^{1\times N}$. 
And N denotes the product of H and W. Similarly, the corresponding result of the fake NTIR 
image is denoted as $V_{fb}^{tl}\in \mathbb{R}^{1\times N}$. 
Then, the loss $\mathcal{L}_{tla}$ can be calculated based on cosine similarity as
\begin{equation}
  \label{L_tba}
  \mathcal{L}_{tla}=\max \left( \theta_{sim}- \frac{V_{ra}^{tl}\cdot V_{fb}^{tl}}{\left \| V_{ra}^{tl} \right \|_{2} \cdot \left \| V_{fb}^{tl} \right \|_{2}},0 \right), 
\end{equation}
where $\theta_{sim}$ denotes the similarity threshold and $\left \| \cdot \right \|_{2} $ 
denotes the L2 norm.

Since the region corresponding to the SOC in the pseudo-labels is noisy, the 
introduction of the above mask-based loss can lead to degradation of edge consistency. 
Considering that structured gradient alignment (SGA) loss \cite{2022-TITS-Luo} can 
encourage the edge consistency of translation, we regularize the geometric consistency of 
corresponding region using SGA loss $\mathcal{L}_{sga}$. Concretely, the edge map of the DC image 
and the gradient map 
of the fake NTIR image after masking the street light and traffic light are denoted 
as $EM_{ra}^{l}$ and $GM_{fb}^{l}$, respectively. Then, the artifact bias correction loss 
can be written as
\begin{equation}
  \label{L_abc}
  \mathcal{L}_{abc}=\mathcal{L}_{sla}+\mathcal{L}_{tla}+ \lambda_{sga} \mathcal{L}_{sga}\left ( GM_{fb}^{l},EM_{ra}^{l} \right ), 
\end{equation}
where $\lambda_{sga}$ denotes the weight of SGA loss, which is set to 0.5 according to 
\cite{2022-TITS-Luo}.

\subsubsection{Color Correction for Image Reconstruction}
Although PearlGAN \cite{2022-TITS-Luo} can effectively reduce the reconstruction error of 
images by combining the smooth L1 loss \cite{2017-CVPR-Zhu} $\mathcal{L}_{sl1}$ and 
SSIM \cite{2004-TIP-Wang} loss $\mathcal{L}_{ssim}$, the reconstruction of SOC regions will 
often show color distortion due to the small percentage of contribution to the total loss, as shown in 
Fig. \ref{fig_bc}(c). To address this issue, with 
reference to \cite{2022-TITS-Luo}, we first define a mask-based image distance 
function $\mathcal{MIDF}\left(\cdot \right)$. Given the original image $I_{ori}\in \mathbb{R}^{C\times H\times W}$, 
the reconstructed image $I_{rec}\in \mathbb{R}^{C\times H\times W}$, and the binary 
mask $M_{b}\in \mathbb{R}^{H\times W}$, the image distance corresponding to the mask 
region can then be expressed as
\begin{equation}
  \label{F_img}
  \begin{split}
    \mathcal{MIDF} \left(M_{b},I_{rec},I_{ori} \right) = &\mathcal{L}_{ssim}\left(M_{b}\odot I_{rec},M_{b}\odot I_{ori} \right) +\\
    &\lambda_{sl1} \mathcal{L}_{sl1}\left(M_{b}\odot I_{rec},M_{b}\odot I_{ori} \right) , 
  \end{split} 
\end{equation}
where $\odot$ denotes element-wise multiplication with channel-wise broadcasting, and 
the weight $\lambda_{sl1}$ is set to 10.0 according to \cite{2017-CVPR-Zhu}. 

Then, we propose a color bias correction loss $\mathcal{L}_{cbc}$ based on the function 
$\mathcal{MIDF}\left(\cdot \right)$ to reduce the color distortion involved in reconstruction process. In 
particular, given the binary mask $M_{SOC}^{A}$ of SOC, the reconstructed 
image $\tilde{x}_{ra}$, and the original image $x_{ra}$, the loss $\mathcal{L}_{cbc}$ can 
then be formulated as
\begin{equation}
  \label{L_cbc}
  \mathcal{L}_{cbc}=\mathcal{MIDF} \left( M_{SOC}^{A},\tilde{x}_{ra},x_{ra} \right ).
\end{equation}
Finally, the total bias correction loss $\mathcal{L}_{bc}$ is the sum of $\mathcal{L}_{abc}$ 
and $\mathcal{L}_{cbc}$. With the introduction of bias correction loss, the SOC 
reconstruction error of fake 
NTIR branch will be significantly reduced, which paves the way for improving the 
colorization performance of the SOC in real NTIR images.
\subsection{OAMix Module}
\begin{figure}[!t]
\centering
\includegraphics[width=3.45in]{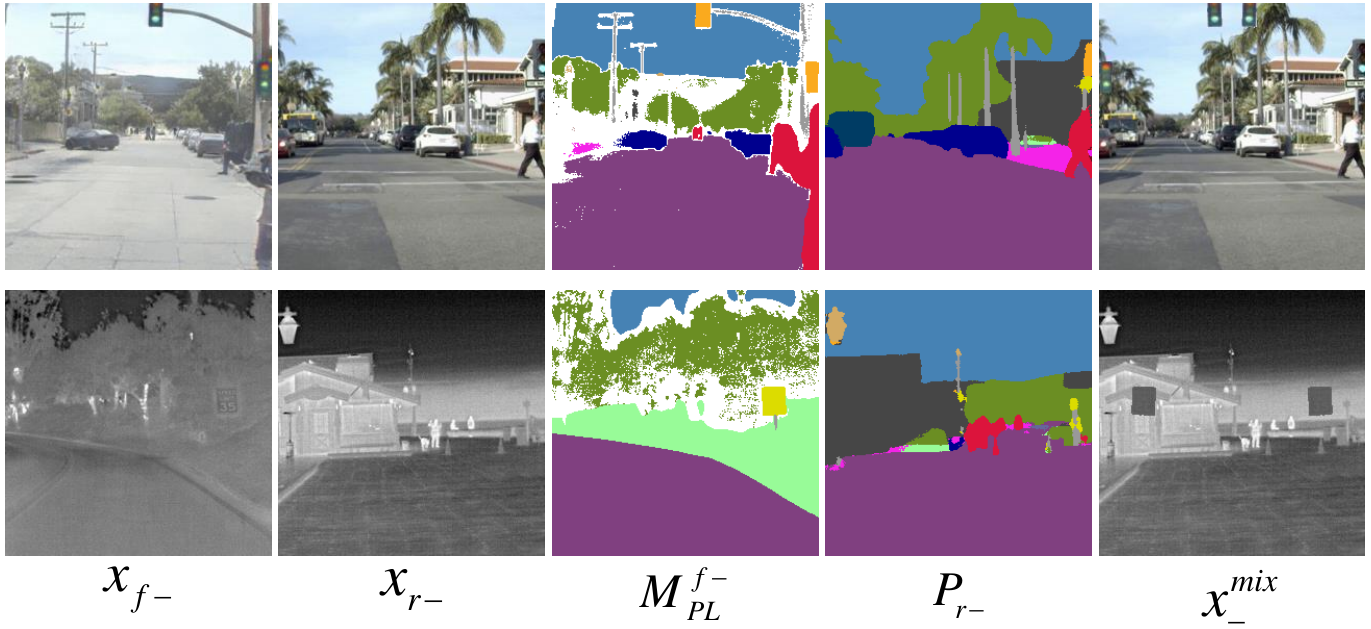}
\caption{Example images of the fusion results in two domains obtained by the OAMix module. 
The first two columns are the input fake and real images, respectively. And the last three 
columns are the pseudo-labels of the fake images, the predicted masks of the real images 
and the fusion results, respectively.}
\label{fig_oamix}
\end{figure}

To generalize the mapping of the SOC in fake NTIR branch to real NTIR, we propose an 
OAMix module and an appearance consistency loss. For the mixing of NTIR images, the 
preservation of the contours of objects is crucial for the recognition of objects due 
to lack of color and texture information. Moreover, occlusion between objects may 
cause unreasonable mixing of DC images, e.g., a motorcycle appears in the face of 
pedestrians. Therefore, unlike previous methods, we propose an OAMix module to 
avoid object occlusion during image mixing. In addition, considering the sparsity of 
SOC samples, the mixed SOC objects contain instances from both the original image and 
its horizontally flipped image, an example image is shown in Fig. \ref{fig_oamix}.

Taking the DC domain as an example, given the fake DC 
image $x_{fa}\in \mathbb{R}^{C\times H\times W}$ and the real DC 
image $x_{ra}\in \mathbb{R}^{C\times H\times W}$, and their semantic 
pseudo-labels $M_{PL}^{B}\in \mathbb{R}^{H\times W}$ and $P_{ra}\in \mathbb{R}^{H\times W}$, 
the OAMix module aims to fuse the SOC in $x_{fa}$ with $x_{ra}$. Specifically, we first 
obtain the binary masks of objects and roads of $x_{ra}$ by $P_{ra}$, denoted 
as $M_{obj}^{A}$ and $M_{road}^{A}$, respectively. Then, given the SOC 
set $\mathcal{C}_{so}$, for the category $c_{i} \in \mathcal{C}_{so}$, we can obtain 
the binary mask $M_{i}^{B}\in \left \{ 0,1 \right \}^{H\times W}$ for the 
category $c_{i}$ of $x_{fa}$. Subsequently, we 
can obtain all the connected regions of $M_{i}^{B}$ based on the connectivity. For the $j_{th}$ 
connected region $M_{i,j}^{B}$ whose area is larger than the threshold, 
if the intersection of $M_{i,j}^{B}$ and $M_{obj}^{A}$ is empty, we determine 
whether it is a subset of the original mixing 
mask $Q^{Ao}\in \left \{ 0,1 \right \}^{H\times W}$ according to the following rules
\begin{equation}
  \label{Q_ij}
  \begin{split}
    Q_{i,j}^{Ao}  =
  \begin{cases}
    M_{i,j}^{B}, & \mbox{if $c_{i}\notin  \mathcal{C}_{veh}   .$} \\
    M_{i,j}^{B}, & \mbox{if $c_{i}\in   \mathcal{C}_{veh}, M_{i,j}^{B}\cap M_{road}^{A}\ne \varnothing  .$} \\
    0, & \mbox{otherwise.}
  \end{cases}
  \end{split} 
\end{equation}
$\mathcal{C}_{veh}$ denotes the set of categories that belong to vehicle. Given that the 
number of $\mathcal{C}_{so}$ is $N_{soc}$, and the number of connected regions corresponding 
to the $i_{th}$ category is $N_{cr}^{i}$, we can obtain the original mixing mask by
\begin{equation}
  \label{Q_AO}
  Q^{Ao}=\sum_{i=1}^{N_{soc} } \sum_{j=1}^{N_{cr}^{i}} Q_{i,j}^{Ao}.
\end{equation}
Similarly, for the result of horizontal flip of $x_{fa}$, denoted as $x_{fa}^{f}$, we can 
obtain the corresponding mixing mask $Q^{Af}$. Next, we can utilize $Q^{Ao}$ and $Q^{Af}$ to 
obtain the mixed object areas
\begin{equation}
  \label{x_obj}
  x_{fa}^{obj} =Q^{Ao}\odot x_{fa} +  \left(Q^{Af}-\left ( Q^{Af}\cap Q^{Ao}\right) \right) \odot x_{fa}^{f}. 
\end{equation}
And the final mixed image can be represented as
\begin{equation}
  \label{x_mix}
  x_{a}^{mix} =x_{fa}^{obj}+ \left ( \mathbf{1}- \left ( Q^{Ao}\cup Q^{Af} \right ) \right ) \odot x_{ra} .
\end{equation}
where $\mathbf{1}$ is a binary mask filled with ones. Next, we can use the image distance 
function $\mathcal{MIDF}\left(\cdot \right)$ of Eq. (\ref{F_img}) to encourage the appearance consistency of SOC 
objects. Let the translated result of image $x_{a}^{mix}$ be denoted as $x_{ab}^{mix}$, 
then the appearance consistency loss of domain A can be expressed as
\begin{equation}
  \label{L_aca}
  \mathcal{L}_{ac}^{a} =\mathcal{MIDF} \left( Q^{Ao},x_{ab}^{mix},x_{rb} \right ).
\end{equation}

Similarly, we can obtain two mixing mask of domain B, denoted as $Q^{Bo}$ and $Q^{Bf}$, 
and the mixed object region, denoted as $x_{fb}^{obj}$. However, due to the temperature 
difference among NTIR images, direct mixing of $x_{fb}^{obj}$ and $x_{rb}$ may produce 
obvious stitching edges, which deviate from the temperature distribution of the current 
scene. Therefore, we propose an adaptive 
luminance adjustment (ALA) strategy to make the mixing of NTIR images more natural. Considering 
the high frequency of road areas in traffic images and their relatively uniform 
temperature distribution, the ALA strategy aims to encourage the average temperature 
of $x_{fb}^{obj}$ to be similar to that of $x_{rb}$ in road areas. Specifically, let 
the mean value of $x_{fb}^{obj}$ be $\mu_{fb}^{obj}$ and the mean value of the road 
region of $x_{rb}$ be $\mu_{rb}^{road}$, then the mixed image of domain B can be represented as
\begin{equation}
  \label{x_mixb}
  x_{b}^{mix} =\frac{\mu_{rb}^{road}}{\mu_{fb}^{obj}} \times x_{fb}^{obj}+ \left ( \mathbf{1}- \left ( Q^{Bo}\cup Q^{Bf} \right ) \right ) \odot x_{rb},
\end{equation}
where the fractional in the first term represents the luminance adjustment factor. 
When the value of the fractional is greater than 1.0, the mixed object 
area $x_{fb}^{obj}$ needs to be 
brightened to adapt to the current scene, and vice versa. Such an 
adaptive sample fusion not only makes the temperature distribution of the mixed 
NTIR images more reasonable, but also reduces the overfitting probability of SOC translation.

For the appearance consistency loss of domain B, we introduce conditional gradient repair 
loss \cite{2022-Arxiv-Luo} $\mathcal{L}_{cgr}$ to reduce the image gradient disappearance 
during translation, in addition to the supervised loss on object regions. Specifically, let 
the translated result of the image $x_{b}^{mix}$ be denoted as $x_{ba}^{mix}$, then the 
appearance consistency loss of domain B can be expressed as
\begin{equation}
  \label{L_acb}
  \begin{split}
    \mathcal{L}_{ac}^{b} = &\mathcal{MIDF}\left( Q^{Bo},x_{ba}^{mix},x_{ra} \right)+\\
    &\mathcal{L}_{cgr}\left(Q_{con}^{B}\odot  x_{ba}^{mix},Q_{con}^{B}\odot x_{rb}\right), 
  \end{split} 
\end{equation}
where $Q_{con}^{B}$ denotes the binary mask of the context in domain B, which can be 
represented as
\begin{equation}
  \label{Q_conb}
  Q_{con}^{B}=\mathbf{1}- \left ( Q^{Bo}\cup Q^{Bf} \right ) .
\end{equation}
Ultimately, the total appearance consistency loss $\mathcal{L}_{ac}$ is the sum 
of $\mathcal{L}_{ac}^{a}$ and $\mathcal{L}_{ac}^{b}$.

With the introduction of the OAMix module and appearance consistency loss, SOC objects 
are required to maintain robust translation while the context changes, which helps reduce 
the context dependence of object translation.
\subsection{Traffic Light Appearance Loss}
Although the problem of context dependence has been moderated as above, how to improve the 
appearance realism of traffic lights in colorization results is under-explored. Under 
nighttime conditions, traffic lights are very common and the glare from them may 
largely interface with the depth estimation 
of their neighboring areas. Hence, reasonable traffic light colorization by the NTIR2DC method 
is beneficial for fine-grained scene perception. Therefore, considering the luminance 
and color, a traffic light appearance loss is proposed to enhance the colorization 
performance of traffic lights.
\subsubsection{Traffic Light Luminance Loss}
Previous NTIR2DC methods \cite{2022-TITS-Luo,2022-Arxiv-Luo} usually change the luminance 
distribution of traffic lights during the translation process, i.e., translate the 
bright regions in the NTIR image into dark regions. This distribution flipping phenomenon 
deviates from the prevalent mapping relationship between temperature and luminance, i.e., 
the temperature of the bright regions of traffic lights is higher than that of their 
neighbors. Therefore, the traffic light luminance loss $\mathcal{L}_{tll}$ is proposed to 
encourage a positive scale mapping of temperature to luminance. 

Specifically, based on 
the mean luminance of traffic lights in the NTIR image $x_{rb}$, we can 
obtain two binary masks $M_{tl}^{br}$ and $M_{tl}^{dr}$ for the traffic light regions, 
indicating the bright and dark regions of traffic lights, respectively. Then, 
we can obtain the lowest luminance $\delta_{tl}^{br}$ corresponding to $M_{tl}^{br}$ and 
the average luminance $\mu_{tl}^{dr}$ corresponding to $M_{tl}^{dr}$ of the fake DC 
image $x_{fa}$. Next, the loss $\mathcal{L}_{tll}$ encourages that the minimum value of 
the translated high-temperature region should be no less than the mean value of the 
low-temperature region, which can be formulated as
\begin{equation}
  \label{L_tll}
  \mathcal{L}_{tll}=\frac{\max \left ( \mu_{tl}^{dr}-\delta_{tl}^{br},0 \right ) }{\delta_{tl}^{br}},
\end{equation}
where the denominator is used for adaptive scaling of the results.
\subsubsection{Traffic Light Color Loss and Color Conversion Module}
Although the colors of traffic lights are not unique, they are closely related to spatial 
locations, i.e., lights that are lit at similar locations usually show same colors. 
Therefore, the traffic light color loss $\mathcal{L}_{tlc}$ aims to guide the model to 
learn the correspondence between the colors and the luminous positions. Among 
the colors of traffic lights, yellow is difficult to be captured by visible cameras due to 
the short luminous time. Thus, the traffic light colors to be recovered in this paper are 
mainly red and green. However, due to the chance of data collection, the data of red and 
green lights are usually unbalanced, which introduces a large bias in color learning. 
In view of this, we creatively propose a color conversion module to realize the 
interconversion of red and green.

Given a traffic light instance in a DC image (RGB format), if its aspect ratio is greater 
than the given threshold, we first flip it vertically and then convert it to 
HSV format. Next, we obtain the binary masks corresponding to the red and green regions, 
denoted as $M_{rr}^{ins}$ and $M_{gr}^{ins}$, respectively, based on the hue values. 
If $M_{rr}^{ins}$ is greater than $M_{gr}^{ins}$, we map the red regions to the green 
regions according to the mapping relationship of hue values. Conversely, the green region 
is mapped to the red region. Ultimately, the mapped HSV format is then converted to RGB 
format. The above processing is integrated into the color conversion module. 
To reduce the learning bias of colors, the random 
vertical flip of traffic light instances is embedded in the OAMix-TIR module, and its 
corresponding label $x_{ra}$ is processed using the color conversion module.

Combined with the color conversion module, the loss $\mathcal{L}_{tlc}$ can guide the model 
to learn the colors of traffic lights more efficiently. Specifically, given the traffic 
light mask $M_{tl}^{B}\in \left \{ 0,1 \right \}^{H\times W}$ of the NTIR image, we first 
divide it into two binary sub-masks---upper and lower, denoted $M_{upr}^{B}$ and $M_{lowr}^{B}$, 
according to the center height of each instance. Similarly, the two sub-masks of the DC 
image $x_{ra}\in \mathbb{R}^{C\times H\times W}$ are denoted 
as $M_{upr}^{A}$ and $M_{lowr}^{A}$, respectively. Then, we use 
the color conversion module to obtain the bright region mask of each instance, which are 
subsequently fused to obtain the bright region mask of $x_{ra}$, denoted as $M_{br}^{A}$. 
For the NTIR image $x_{rb}$, the corresponding bright region mask of traffic light is the 
same as $M_{tl}^{br}$ mentioned in the previous subsection, denoted as $M_{br}^{B}$. Next, 
for the traffic light region of $x_{ra}$, the average color feature of the bright 
region (e.g., red light) in its upper sub-masks, 
denoted as $F_{ub}^{ra}\in \mathbb{R}^{C\times 1\times 1}$, can be expressed as
\begin{equation}
  \label{F_ub}
  F_{ub}^{ra}= \frac{H\times W}{N_{ub}^{A}  } \times \mathcal{GAP} \left ( M_{upr}^{A}\odot M_{br}^{A}\odot x_{ra}\right ), 
\end{equation}
where $N_{ub}^{A}$ denotes the sum of the binary masks $M_{upr}^{A}\odot M_{br}^{A}$, and 
$\mathcal{GAP}\left( \cdot\right)$ denotes the global average pooling operation. Similarly, 
we can obtain the average color feature $F_{lb}^{ra}$ for the bright 
region (e.g., green light) in the lower sub-masks, and the upper and lower features corresponding 
to the fake DC image $x_{fa}$, denoted as $F_{ub}^{fa}$ and $F_{lb}^{fa}$, respectively. 
Then, the color distance of the upper sub-masks can be presented as
\begin{equation}
  \label{d_col}
  d_{uu}^{fr} =\sqrt{\sum_{k=1}^{C}\left( \left( F_{ub}^{fa}- F_{ub}^{ra} \right )^{2}\right )_{k}}. 
\end{equation}
Similarly, we can obtain the color distance $d_{ll}^{fr}$ for 
the lower half. Due to the small difference between red and green in the root mean square 
distance and the sparsity of green lights on traffic images, an appropriate weight 
on $d_{ll}^{fr}$ is necessary for the learning of green lights. To encourage that the 
intra-class distance (i.e., $d_{ll}^{fr}$) of colors should be smaller 
than the inter-class distance (i.e., the distance between $F_{lb}^{fa}$ and $F_{ub}^{ra}$, 
denoted as $d_{lu}^{fr}$), we set the weighting factor as
\begin{equation}
  \label{beta_w}
  \beta _{w}=\frac{1}{\min \left( d_{ll}^{fr},d_{lu}^{fr} \right )+ \tau},
\end{equation}
where $\tau$ is a constant to prevent the denominator from being zero as well as to 
control the range of values, which is empirically set to 0.05. Ultimately, the complete 
color loss is expressed as
\begin{equation}
  \label{L_tlc}
  \mathcal{L}_{tlc}=d_{uu}^{fr}+\beta _{w}\times d_{ll}^{fr}.
\end{equation}
Considering both luminance and color, the traffic light appearance loss $\mathcal{L}_{tla}$ 
is the sum of loss $\mathcal{L}_{tll}$ and $\mathcal{L}_{tlc}$.

\subsection{Dual Feedback Learning Strategy}
Although the OAMix module and traffic light appearance loss are beneficial for feature 
learning of SOC, the boosting is limited due to the low  and relatively fixed learning 
frequency. In contrast to fixed learning schedules, humans use feedback to enhance 
the systematicity of learning strategies for skill 
acquisition \cite{2005-LI-Vollmeyer,2014-BBR-Luft}. Inspired by such a feedback-based 
learning scheme, we propose a dual feedback learning strategy to dynamically adjust the 
learning frequency of SOC samples based on the current learning state. 

Let the SOC sample sets of domain A and domain B be defined as $X_{soc}^{A}$ 
and $X_{soc}^{B}$, respectively. 
Then, we evaluate the learning status of the model for SOC by comparing the 
appearance consistency loss of SOC with the global reconstruction loss, which determines 
whether the learning of SOC samples requires enhancement. Taking domain A as an example, given 
the values of the appearance consistency loss (i.e., Eq. (\ref{L_aca})) for SOC and the global 
reconstruction loss (i.e., $\mathcal{MIDF}\left (\mathbf{1},\tilde{x}_{ra},x_{ra} \right)$) 
at moment $t-1$, defined as $Z_{As}^{\left(t-1 \right)}$ and $Z_{Ag}^{\left(t-1 \right)}$, 
respectively, the image sampled at moment $t$ according to the dual feedback learning 
strategy can be expressed as
\begin{equation}
  \label{dfb}
  \begin{split}
    x_{ra}^{\left ( t \right ) }  \in 
  \begin{cases}
    X_{soc}^{A}, & \mbox{if $Z_{As}^{\left(t-1 \right)}> Z_{Ag}^{\left(t-1 \right)}.$} \\
    X^{A}, & \mbox{otherwise.}
  \end{cases}
  \end{split} 
\end{equation}
When $Z_{As}^{\left(t-1 \right)}$ is greater than $Z_{Ag}^{\left(t-1 \right)}$, the appearance 
learning for SOC is below the mean, which indicates that the model needs to enhance 
the learning frequency for SOC samples. Conversely, the model does not require changes to 
the original learning plan. The sampling strategy for domain B is also similar to Eq. (\ref{dfb}). 
By dynamically adjusting the learning strategy through the dual feedback of 
both domains, the model can not only allocate learning resources more rationally (e.g., 
reduce redundant learning of simple samples), but also learn the features of the SOC efficiently. 

\subsection{Objective Function}
In summary, the overall objective function of the proposed FoalGAN can be expressed as
\begin{equation}
  \label{L_all}
  \mathcal{L}_{all} =\mathcal{L}_{base}+\mathcal{L}_{bc}+\mathcal{L}_{ac}+\mathcal{L}_{tla}.
\end{equation}
\section{Experiments}
In this section, we first introduce the datasets and evaluation metrics for the NTIR2DC 
task. Subsequently, we explain the annotation of the Brno dataset and the implementation
details of FoalGAN. Then, experimental results on the FLIR and Brno datasets are presented. 
Next, we perform an ablation analysis to verify the validity of the proposed modules, 
losses, and strategies. At last, some discussions of the experimental results are presented.
\subsection{Datasets and Evaluation Metrics}
\subsubsection{Datasets}
Due to the poor quality of the NTIR images in the KAIST \cite{2015-CVPR-Hwang} 
dataset (e.g., the SOC regions are too blurred to be detected), we choose the FLIR and Brno 
datasets to evaluate the translation performance of the SOC. The FLIR Thermal Starter 
Dataset \cite{2019-FLIR-FA} provides an annotated TIR image set and a non-annotated RGB 
image set for training and validating object detection models. Through the same data split 
as in \cite{2022-TITS-Luo}, we finally obtain 5447 DC images and 2899 NTIR images for training,
while an additional 490 NTIR images are used for testing.

The Brno Urban Dataset \cite{2020-ICRA-Ligocki} records data from multiple sensors in the city of Brno for the study of autonomous driving 
tasks. The data were collected in sunny, cloudy and rainy environments. Three videos of sunny 
days in the DC domain, and two videos of cloudy days and one video of rainy days in the NTIR 
domain were selected as training data. An additional cloudy video and a rainy video in the 
NTIR domain were selected as test data. To eliminate redundant data at adjacent times, we 
re-sampled the videos at one frame per second, and finally obtained 2574 DC images and 2383 
NTIR images for model training. Similarly, we re-sampled the tested NTIR videos at an interval 
of 60 frames, and finally selected 500 NTIR images randomly as the test 
set\footnote{See \url{https://github.com/FuyaLuo/FoalGAN/} for detailed sample selection.}.

In order to remove the black areas on both sides in some images, according to 
\cite{2022-TITS-Luo}, we first resize the training images to a resolution 
of $500\times 400$, and then the $360\times 288$ resolution images obtained by center 
cropping are used as training data.
\subsubsection{Evaluation Metrics}
To evaluate the performance for image content 
preservation at each level, we conduct experiments on three 
vision tasks: semantic segmentation, object detection, and edge preservation.

Intersection-over-Union (IoU) \cite{2015-IJCV-Everingham} is a widely used metric in 
semantic segmentation tasks. The mean value of IoU 
for all classes, denoted as mIoU, is adopted to evaluate the semantic consistency of NTIR 
image colorization methods.

Average precision (AP) \cite{2010-IJCV-Everingham} denotes the average detection 
precision of the object detection model under different recalls. The mean value of AP for 
all categories, defined as mAP, is selected as an overall evaluation metric.

APCE \cite{2022-TITS-Luo} is the average precision of Canny edges under multi-threshold 
conditions, and is employed to evaluate the edge preservation performance of the NTIR2DC model.

\begin{figure}[!t]
\centering
\includegraphics[width=3.45in]{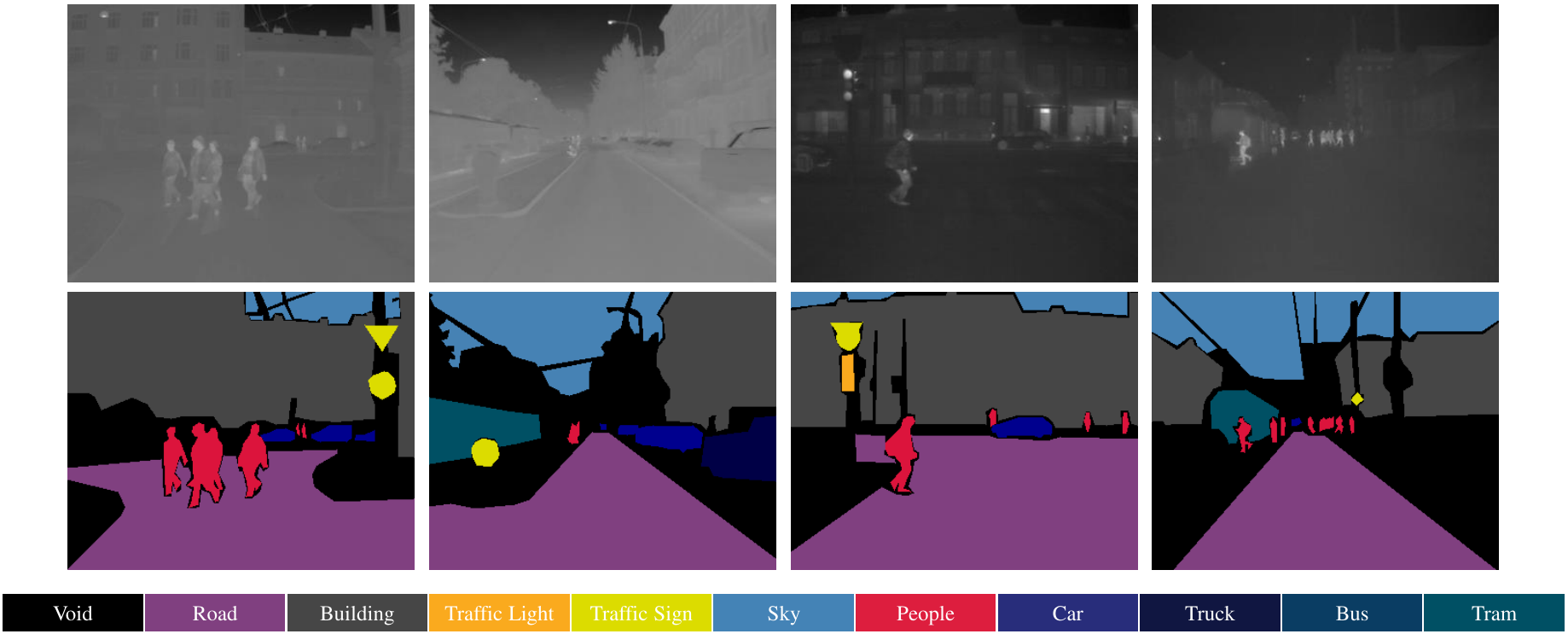}
\caption{Examples of NTIR images from the Brno dataset (top row) and our annotated 
segmentation masks (bottom row).}
\label{fig_ann}
\end{figure}

\subsection{Image Annotation of the Brno Dataset}
To evaluate the semantic preservation performance of the NTIR2DC task, we selected and 
annotated a subset of Brno 
dataset with their pixel-level category labels. Considering the 
diversity of weather, 500 NTIR images in cloudy and rainy environments were selected for 
annotation. Due to the low contrast and ambiguous boundaries of NTIR images, as shown in 
Fig. \ref{fig_ann}, we define ten categories and use the 
LabelMe\footnote{\url{https://github.com/CSAILVision/LabelMeAnnotationTool}.} toolbox to 
annotate only their identified corresponding regions. The labeled categories are road, 
building, traffic light, traffic sign, sky, people, car, truck, bus, and tram.
\subsection{Experimental Settings and Implementation Details}
We compare FoalGAN with other NTIR2DC methods such as PearlGAN \cite{2022-TITS-Luo} 
and MornGAN \cite{2022-Arxiv-Luo}, as well as some low-light enhancement methods (e.g., 
ToDayGAN \cite{2019-ICRA-Anoosheh}) and prevalent I2I 
translation methods (e.g., CycleGAN \cite{2017-CVPR-Zhu}, UNIT \cite{2017-NIPS-Liu} and 
DRIT++ \cite{2020-IJCV-Lee}). We follow the instructions of these methods in order to establish a 
fair setting for comparison.

FoalGAN is implemented using PyTorch. We train the model using the 
Adam \cite{2014-Arxiv-Kingma} optimizer 
with $\left(\beta_{1},\beta_{2}\right) =\left( 0.5,0.999 \right)$ on NVIDIA RTX 3090 GPUs. The
batch size is set to 1 for all experiments. The learning rate of the whole training process 
is maintained at 0.0002. The total number of training epochs for the FLIR and Brno datasets 
are 100 and 160, respectively. In Eq. (\ref{L_sba}), the temperature threshold $\theta_{tem}$ 
is empirically set to 0.25 based on the brightness difference between categories in the NTIR 
images. Referring to the gradient threshold in SGA \cite{2022-TITS-Luo} loss, the similarity 
threshold $\theta_{sim}$ in Eq. (\ref{L_tba}) is set to 0.8. In Section III, 
the SOC set $\mathcal{C}_{so}$ includes three 
categories: traffic light, traffic sign, and motorcycle. For data augmentation, we 
flip the images 
horizontally with a probability of 0.5, and randomly crop them to the size of $256\times 256$. 
The number of parameters of our model is about 46.7 MB, and the inference speed on an NVIDIA 
RTX 3090 GPU is about 0.01 seconds for an input image with a resolution of $360\times288$ pixels.

Due to the lack of category annotation of NTIR images in the training set, we divided the 
training process into two phases: a pre-training phase and a feedback learning 
phase\footnote{See \url{https://github.com/FuyaLuo/FoalGAN/} for implementation 
details.}. In the 
pre-training phase, the model introduces bias correction for learning based on MornGAN to 
determine the SOC sample sets $X_{soc}^{A}$ and $X_{soc}^{B}$, and the training process is 
consistent with MornGAN \cite{2022-Arxiv-Luo}. In the feedback learning phase, the complete 
model dynamically adjusts the learning frequency of the SOC samples based on the current 
learning state, with a total number of iterations of approximately 110K.

Due to the lack of pixel-level annotations on the FLIR and Brno datasets, we evaluate the 
semantic segmentation performance of translated images using a scene parsing 
model HMSANet \cite{2020-Arxiv-Tao} trained on Cityscape \cite{2016-CVPR-Cordts}, which considers both the 
feature plausibility and semantic consistency of colorization. Similarly, to measure the 
naturalness of object features, we utilize YOLOv7 \cite{2022-Arxiv-Wang}, which is 
trained on the MS COCO \cite{2014-ECCV-Lin} dataset as the evaluation model for object 
detection on the translated images.
\subsection{Experiments on the FLIR Dataset}

\begin{figure}[!t]
\centering
\includegraphics[width=3.45in]{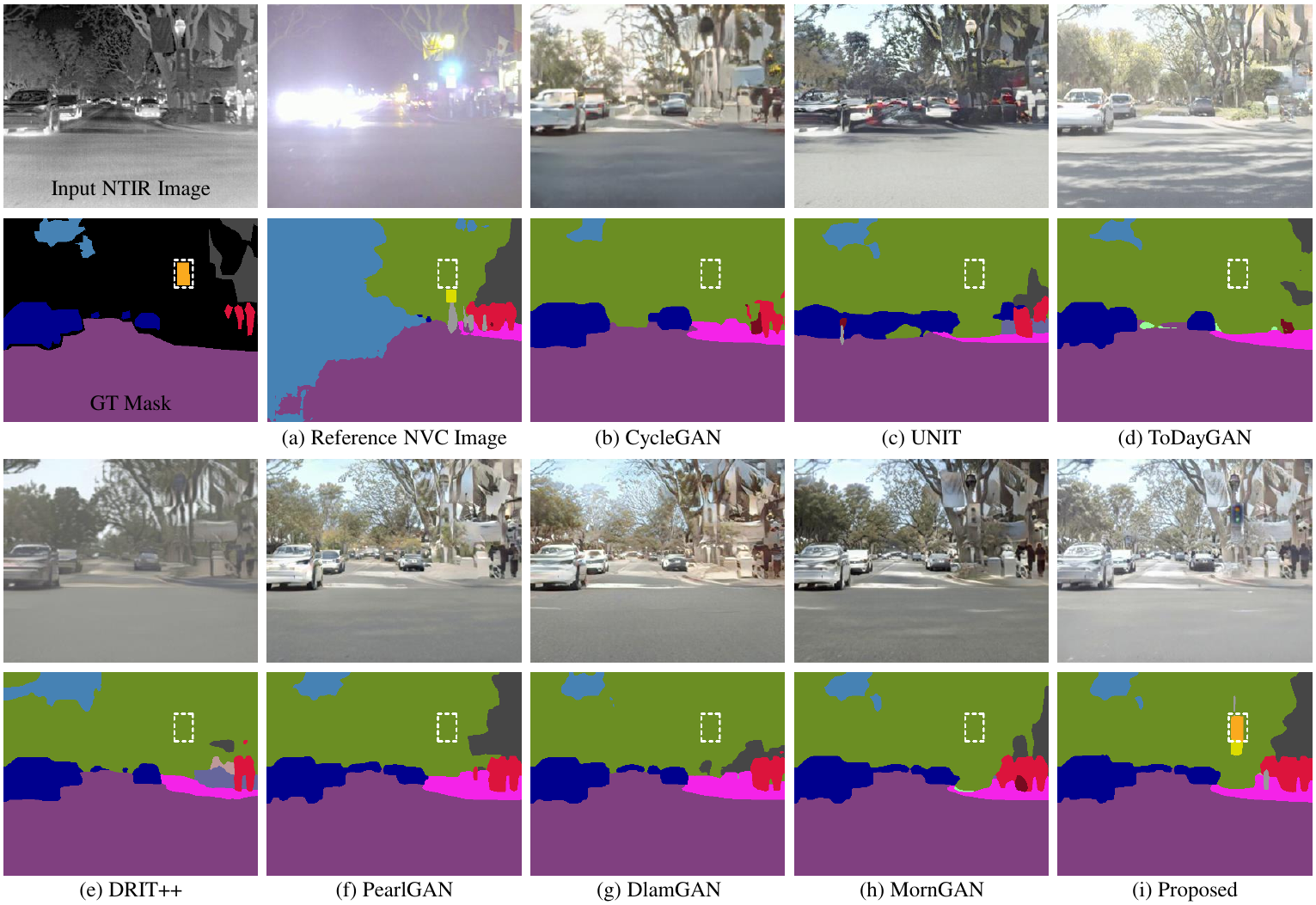}
\caption{The visual comparison of translation (the first row) and segmentation results (the second row) 
for different methods on the FLIR dataset. The areas in the white dotted boxes deserve attention.}
\label{fig_seg_flir}
\end{figure}

\subsubsection{Semantic Segmentation}
The translation results and corresponding segmentation outputs of various methods
on the FLIR dataset are presented in Fig. \ref{fig_seg_flir}. Column (a) represents the reference nighttime 
visible color (NVC) image and its semantic segmentation. Due to the blurring of features, 
the segmentation model HMSANet \cite{2020-Arxiv-Tao} is unable to identify traffic lights and cars in the glare region, 
which poses a great challenge to the visible-based scene perception system. All 
the compared I2I translation methods 
cannot generate plausible traffic lights, as shown in the white dashed box area. In 
contrast, the proposed model can recognize traffic lights in NTIR images and achieve 
reasonable colorization, which facilitates nighttime scene understanding. Moreover, 
directly below the white dashed box, all methods except ours fail to generate traffic 
signs recognized by the segmentation model, which demonstrates the superiority of FoalGAN 
in colorization of small objects.

\begin{table*}[htbp]
  \centering
  \caption{Semantic segmentation performance (IoU) on the translated images by different 
  translation methods on the FLIR dataset.}
    \begin{tabular}{cccccccccccc}
      \toprule
          & Road  & Building & Sky   & Person & Car   & Truck & Bus   & Traffic Light & Traffic Sign & Motorcycle & mIoU \\ \hline
    Reference NVC image & 95.2  & 53.7  & 1.0   & 40.3  & 56.6  & 2.5   & 70.1  & 9.4   & 5.2   & 0.0   & 33.4  \\
    CycleGAN \cite{2017-CVPR-Zhu} & 97.2  & 19.6  & 89.4  & 67.1  & 79.3  & 0.3   & 0.8   & 1.5   & 0.0   & 0.0   & 35.5  \\
    UNIT \cite{2017-NIPS-Liu}  & 96.3  & 48.3  & 92.5  & 59.5  & 63.7  & 0.6   & \textbf{49.4} & 5.9   & 0.0   & 12.4  & 42.9  \\
    ToDayGAN \cite{2019-ICRA-Anoosheh} & 97.0  & 42.3  & 83.2  & 56.3  & 76.5  & 0.0   & 6.5   & 1.4   & 0.0   & 2.5   & 36.6  \\
    DRIT++ \cite{2020-IJCV-Lee} & 98.2  & 16.1  & 75.3  & 38.3  & 79.4  & 0.0   & 3.2   & 2.6   & 0.0   & 0.0   & 31.3  \\
    PearlGAN \cite{2022-TITS-Luo} & 98.6  & 71.0  & 95.1  & 84.3  & 89.1  & 1.5   & 0.0   & 4.4   & 0.0   & 12.2  & 45.6  \\
    DlamGAN \cite{2021-ICIG-Luo} & 97.4  & 67.2  & 94.0  & 74.0  & 89.1  & 2.1   & 0.3   & 6.3   & 0.1   & 36.2  & 46.7  \\
    MornGAN \cite{2022-Arxiv-Luo} & \textbf{98.7} & \textbf{84.0} & \textbf{96.8} & \textbf{94.5} & 95.6  & 1.2   & 22.8  & 22.0  & 0.7   & 40.4  & 55.7  \\
    Proposed & \textbf{98.7} & 82.7  & 96.1  & 94.1  & \textbf{96.3} & \textbf{3.2} & 22.2  & \textbf{57.2} & \textbf{4.0} & \textbf{56.6} & \textbf{61.1} \\
    \bottomrule  
  \end{tabular}%
  \label{tab_flir_seg}%
\end{table*}%

Table \ref{tab_flir_seg} reports a quantitative comparison of the semantic consistency performance on the 
FLIR dataset. The proposed method achieves comparable performance with MornGAN in semantic 
preservation of large sample categories (i.e., road, building, sky, pedestrian, and car). 
And all the compared methods have poor translation performance 
for small sample categories (i.e., truck, bus, traffic light, traffic sign, and motorcycle) 
due to few available samples and diverse colors. However, 
benefiting from the feedback-based appearance learning scheme, the proposed method 
significantly outperforms other methods in terms of semantic consistency on 
SOC (i.e., traffic light, traffic sign, and motorcycle). Overall, 
the proposed method leads other methods in terms of scene layout maintenance by a 
significant margin (i.e., at least 5.4\%).

\subsubsection{Object Detection}
\begin{figure}[!t]
\centering
\includegraphics[width=3.45in]{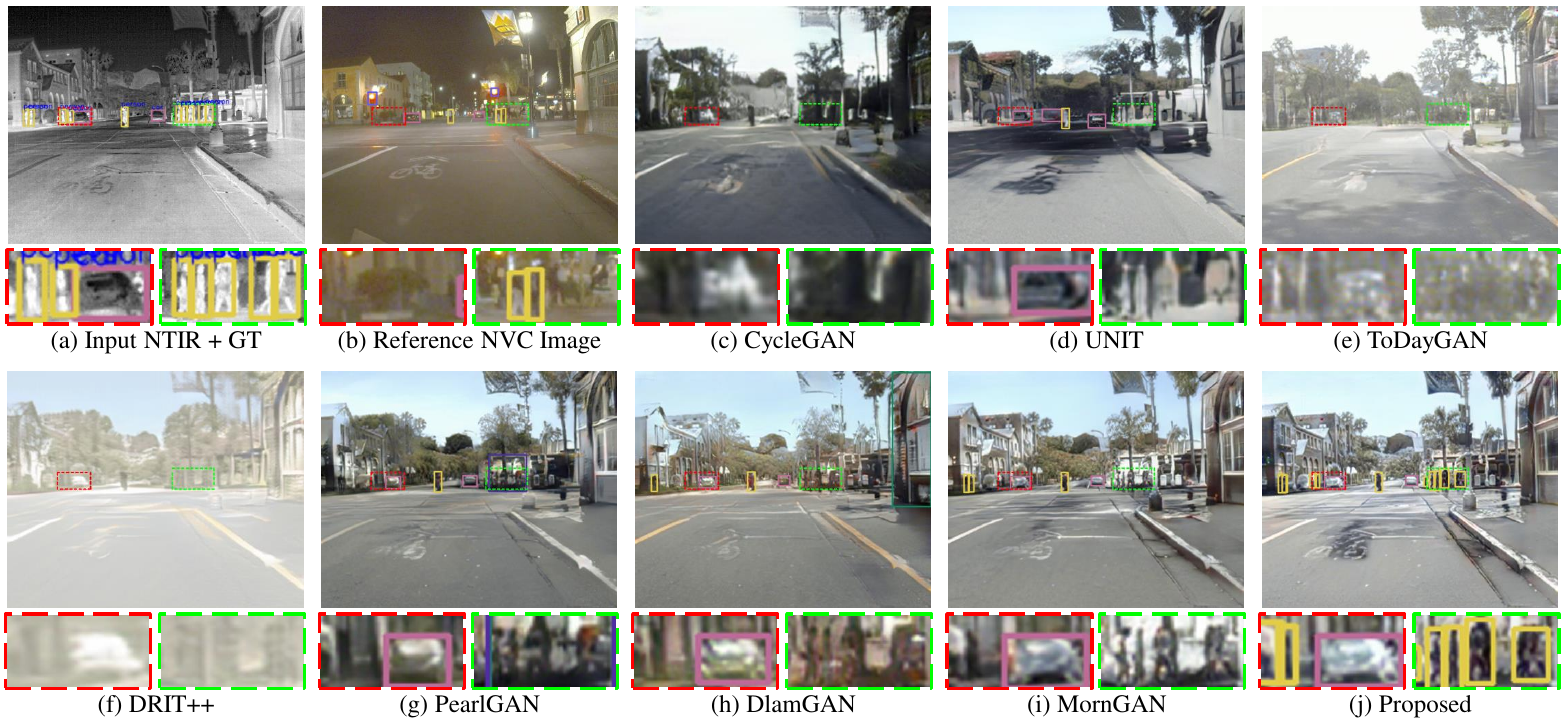}
\caption{Visual comparison of detection results on the FLIR dataset by YOLOv7 
model \cite{2022-Arxiv-Wang}. The parts covered by red and 
green dashed boxes show the enlarged patches in the corresponding images. Colors in the 
detection results that do not intersect with GT represent undefined categories of the FLIR 
dataset as identified by the detector.}
\label{fig_det_flir}
\end{figure}

Fig. \ref{fig_det_flir} shows the qualitative translation and detection result comparisons, wherein the 
second row is the zoomed-in image of the corresponding area. As shown in the dashed box, 
all compared approaches fail to generate plausible pedestrians. Moreover, CycleGAN, 
ToDayGAN and DRIT++ are unable to realistically translate cars. On the contrary, the 
proposed FoalGAN can not only achieve reasonable colorization for small-sized cars, but also 
maintain the complete pedestrian structure, which illustrates the superiority of the 
proposed method for small object preservation. In addition, more pedestrians are detected 
from our translated images than from the NVC images by the object detection model of YOLOv7 \cite{2022-Arxiv-Wang}.

\begin{table}[htbp]
  \centering
  \caption{Object detection performance (AP) on the translated images by different 
  translation methods on the FLIR dataset, computed at a single IoU of 0.50.}
    \begin{tabular}{ccccc} \toprule
          & Person & Bicycle & Car   & mAP \\ \hline
    Reference NVC image & 8.3   & 2.1   & 10.9  & 7.1  \\
    CycleGAN \cite{2017-CVPR-Zhu} & 21.0  & 3.2   & 38.4  & 20.9  \\
    UNIT \cite{2017-NIPS-Liu}  & 17.5  & 11.2  & 20.5  & 16.4  \\
    ToDayGAN \cite{2019-ICRA-Anoosheh} & 25.7  & 1.9   & 53.6  & 27.1  \\
    DRIT++ \cite{2020-IJCV-Lee} & 20.6  & 2.1   & 48.5  & 23.8  \\
    PearlGAN \cite{2022-TITS-Luo} & 59.3  & 25.1  & 75.4  & 53.3  \\
    DlamGAN \cite{2021-ICIG-Luo} & 53.4  & 18.6  & 72.3  & 48.1  \\
    MornGAN \cite{2022-Arxiv-Luo} & 82.7  & 29.7  & 82.9  & 65.1  \\
    Proposed & \textbf{84.6} & \textbf{34.7} & \textbf{85.6} & \textbf{68.3} \\
    \bottomrule  
  \end{tabular}%
  \label{tab_flir_det}%
\end{table}%

As the bounding-box annotation of the FLIR dataset covers only three categories (i.e., 
pedestrian, bicycle, and car), quantitative comparison of various colorization
results for object detection is shown in Table \ref{tab_flir_det}. The proposed method consistently 
outperforms other methods in terms of object preservation for all classes. It is worth 
mentioning that for the retention of challenging bicycles, the proposed FoalGAN still 
outperforms the second-ranked MornGAN by 5\%.

\subsubsection{Edge Preservation}
\begin{figure}[!t]
\centering
\includegraphics[width=3.45in]{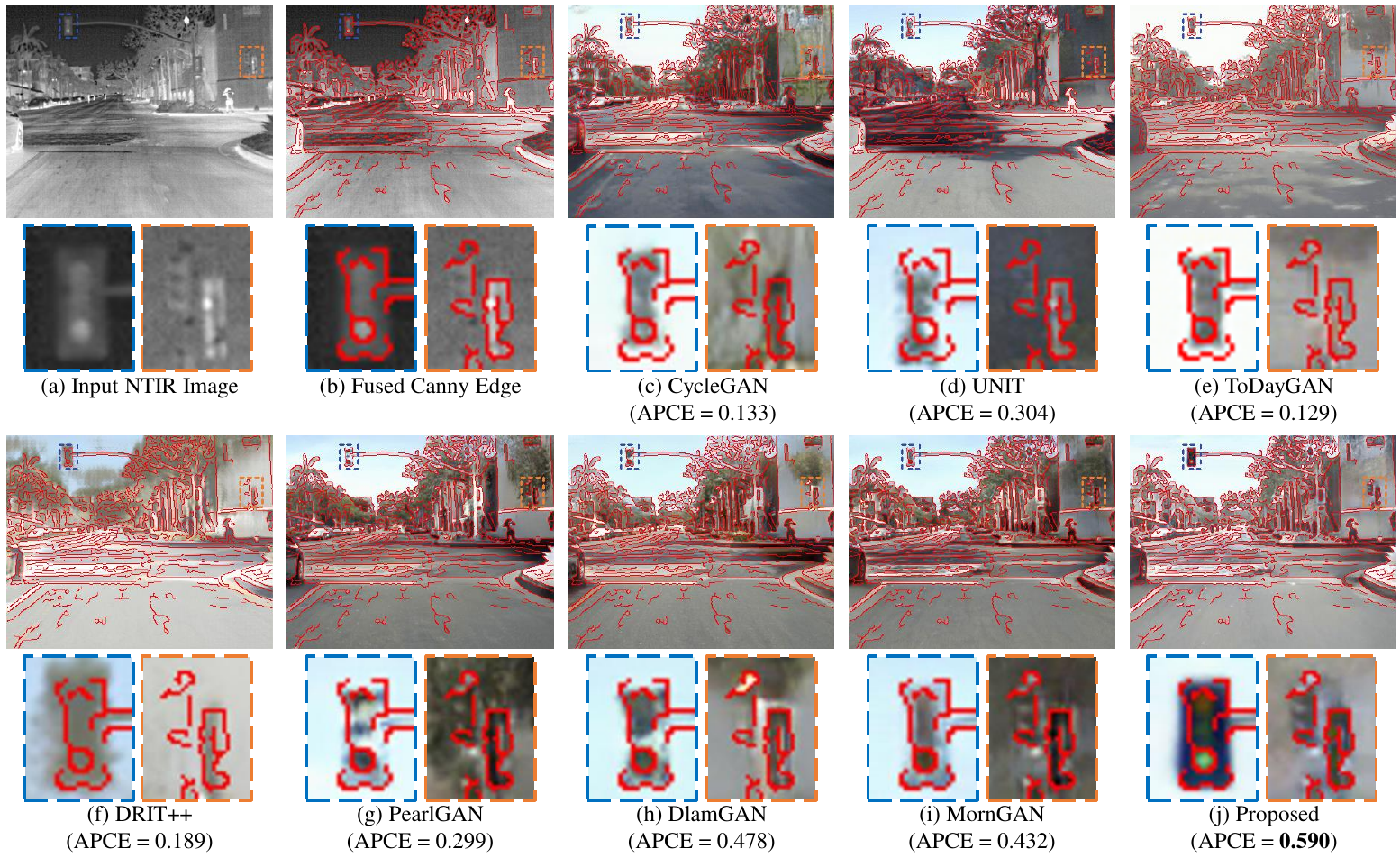}
\caption{Visual comparison of geometric consistency on the FLIR dataset. The second row 
shows the enlarged results of the corresponding regions after fusion with the edges. 
The edges in red are extracted by the Canny detector from the input NTIR image.}
\label{fig_edge_flir}
\end{figure}

In Fig. \ref{fig_edge_flir}, we qualitatively compare the ability of edge preservation 
of different I2I translation 
methods, and the second row shows the zoomed-in patches of the corresponding area fused 
with the edges of the original NTIR image. As shown in the blue dashed box, the edges in 
the DRIT++ results are outwardly expanded relative to the edges of traffic light in 
the NTIR image, while the edges of the other compared methods are inwardly shrinked. On 
the contrary, the edges of the proposed method are able to fit perfectly with the 
original edges, even though the intensity of edges on the left side of traffic 
light is extremely small. Similarly, as shown in the orange dashed box, all methods 
except ours fail to preserve the left edge of the traffic light, which illustrates the 
superiority of the proposed method in terms of structure preservation of small objects.

\begin{figure}[!t]
\centering
\includegraphics[width=3.45in]{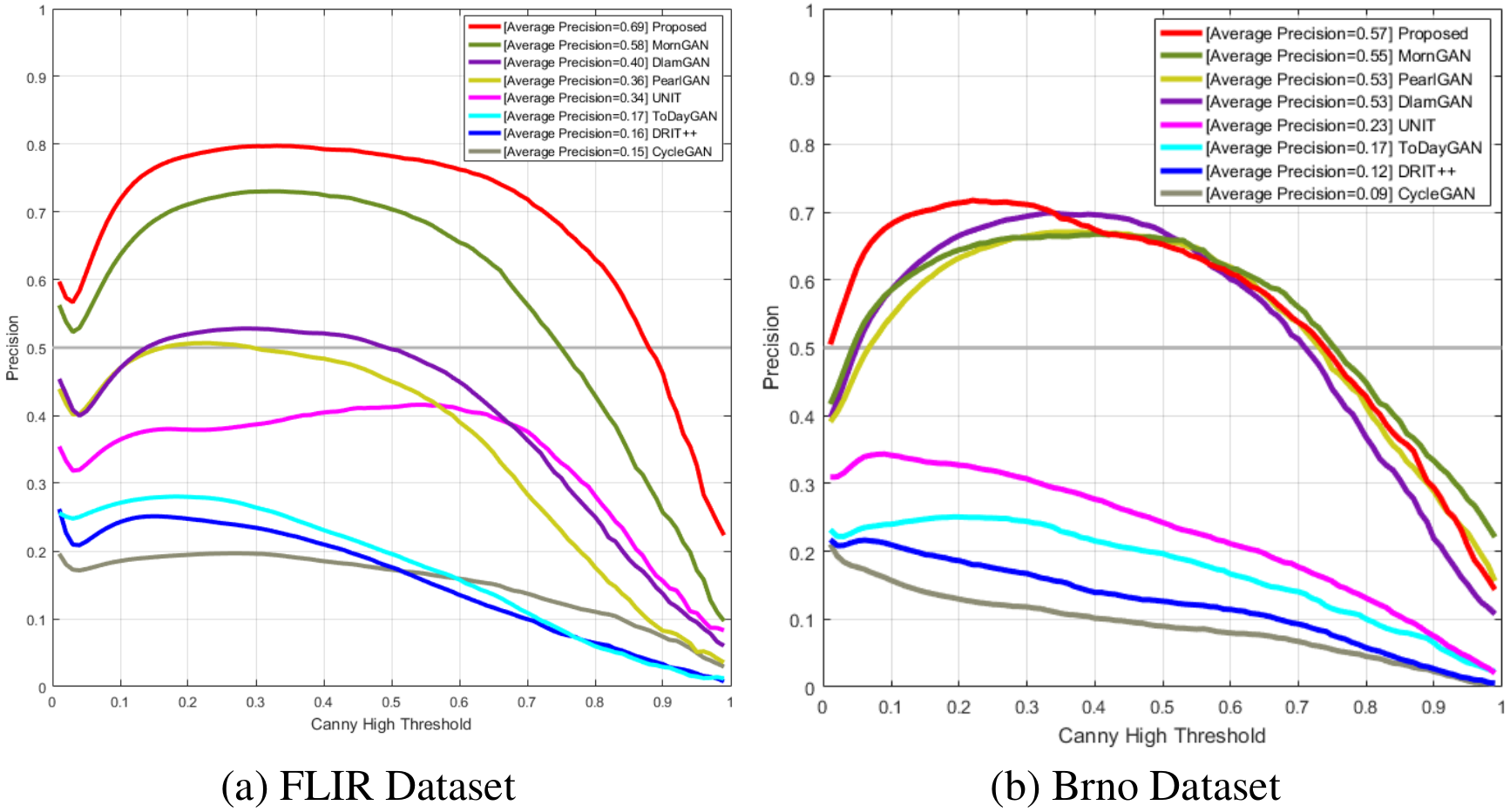}
\caption{APCE results of different translation methods on the FLIR and Brno datasets.}
\label{fig_apce}
\end{figure}

As the Canny edges in the Fig. \ref{fig_edge_flir} are only the results of a fixed threshold, we exploit 
the APCE metric, which covers multiple thresholds to comprehensively evaluate the edge 
consistency performance, as shown in Fig. \ref{fig_apce} (a). We can find that the proposed method 
dramatically outperforms other methods in edge consistency at all thresholds and is 
far superior to the second ranked MornGAN by 11\%.

\subsection{Experiments on the Brno Dataset}
Different from FLIR \cite{2019-FLIR-FA}, Brno \cite{2020-ICRA-Ligocki} is a more 
challenging dataset with NTIR images under multiple weather conditions.
\subsubsection{Semantic Segmentation}
\begin{figure}[!t]
\centering
\includegraphics[width=3.45in]{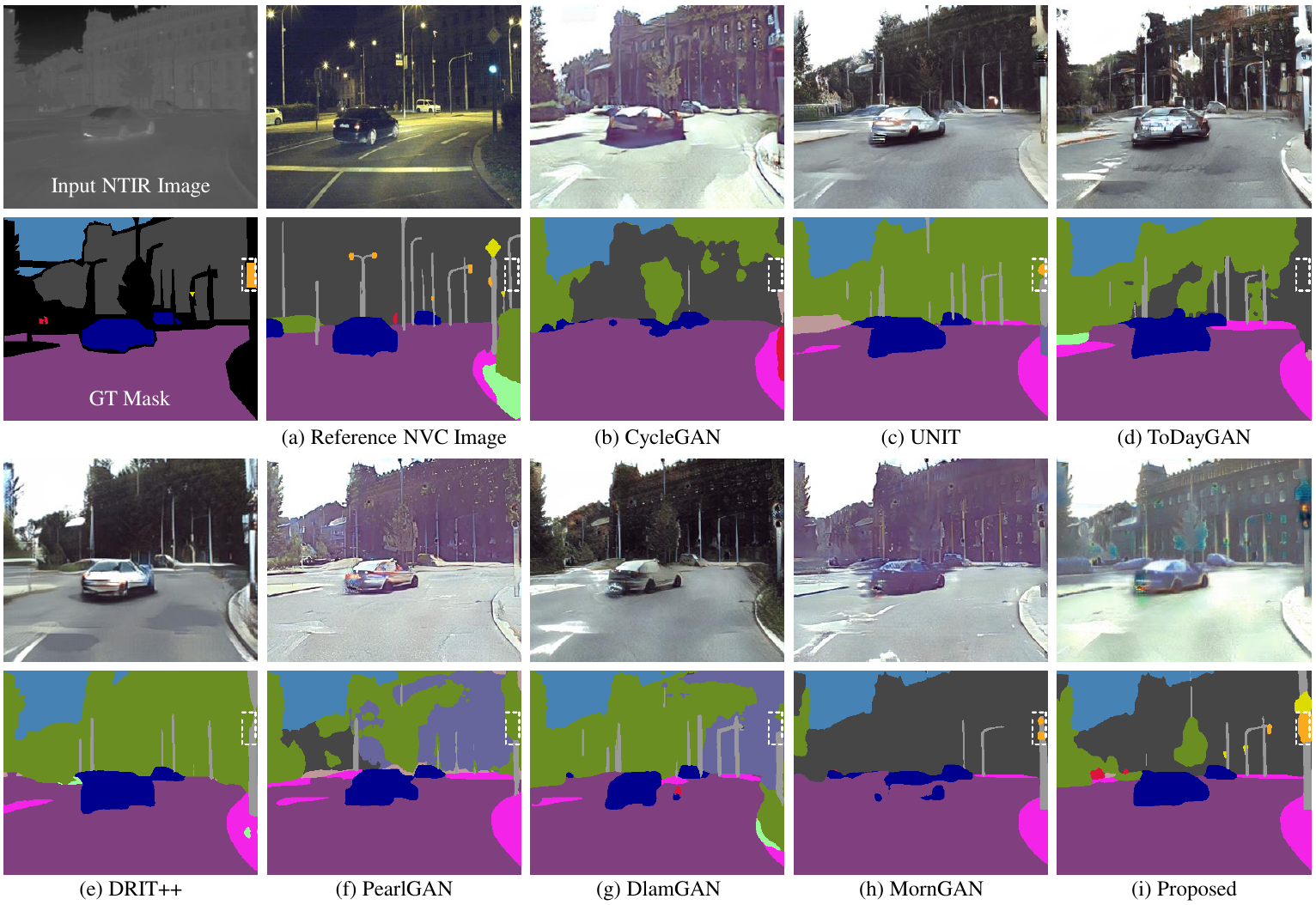}
\caption{The visual comparison of translation (the first row) and segmentation performance 
(the second row) of different methods on the Brno dataset. The areas in white dotted boxes 
deserve attention.}
\label{fig_seg_brno}
\end{figure}

Fig. \ref{fig_seg_brno} presents the translated results and the segmentation outputs of various methods on 
the Brno dataset. As shown in the white dashed box, UNIT and MornGAN can generate some of 
the traffic light features, while none of the other comparison methods can generate 
plausible traffic lights. Unlike other methods, the proposed FoalGAN can retain the 
complete traffic light structure to facilitate the recognition of segmentation models. 
In addition, our translation images are more beneficial for the detection of pedestrians 
on the left side compared with the original NVC images.

\begin{table*}[htbp]
  \centering
  \caption{Semantic segmentation performance (IoU) on the translated images by different 
  translation methods on the Brno dataset.}
    \begin{tabular}{ccccccccccc}
      \toprule
          & Road  & Building & Sky   & Person & Car   & Bus   & Tram  & Traffic Light & Traffic Sign & mIoU \\ \hline
    Reference NVC image & 81.2  & 53.2  & 0.0   & 10.8  & 43.5  & 0.0   & 25.7  & 3.3   & 3.6   & 24.6  \\
    CycleGAN \cite{2017-CVPR-Zhu} & 82.2  & 40.3  & 77.8  & 9.0   & 53.4  & 0.5   & 6.9   & 3.4   & 2.5   & 30.7  \\
    UNIT \cite{2017-NIPS-Liu}  & 91.0  & 53.3  & 85.8  & 21.4  & 68.7  & 6.8   & 13.2  & 9.8   & 3.8   & 39.3  \\
    ToDayGAN \cite{2019-ICRA-Anoosheh} & \textbf{93.9} & 48.7  & 67.5  & 29.4  & 65.0  & 6.9   & 17.5  & 7.3   & 1.6   & 37.5  \\
    DRIT++ \cite{2020-IJCV-Lee} & 93.4  & 55.1  & 74.0  & 19.4  & 68.6  & 0.6   & 12.9  & 2.4   & 2.9   & 36.6  \\
    PearlGAN \cite{2022-TITS-Luo} & 82.1  & 52.4  & \textbf{92.3} & 44.0  & 50.5  & 5.1   & 8.5   & 25.8  & 3.8   & 40.5  \\
    DlamGAN \cite{2021-ICIG-Luo} & 82.2  & 58.0  & 91.9  & 44.7  & 58.0  & \textbf{11.6} & 9.6   & 37.2  & 2.8   & 44.0  \\
    MornGAN \cite{2022-Arxiv-Luo} & 85.7  & 88.8  & 84.6  & 54.4  & 65.0  & 1.4   & 13.6  & 41.3  & 5.3   & 48.9  \\
    Proposed & 92.6  & \textbf{90.4} & 86.5  & \textbf{59.3} & \textbf{72.1} & 6.4   & \textbf{37.0} & \textbf{68.8} & \textbf{28.0} & \textbf{60.1} \\
    \bottomrule  
  \end{tabular}%
  \label{tab_brno_seg}%
\end{table*}%

Furthermore, a quantitative comparison of the semantic preservation performance of various I2I 
translation methods is shown in Table \ref{tab_brno_seg}. The proposed approach achieves the best results in 
terms of semantic preservation for most categories except three (i.e., road, sky, and bus). 
Among them, FoalGAN has a remarkable advantage (i.e., at least 20\%) over other methods in 
the translation of traffic lights and traffic signs, which illustrates the superiority of the 
proposed method for appearance learning of SOC. In general, FoalGAN has a substantial 
lead (i.e., at least 12.2\%) in semantic consistency of colorization relative to other methods.
\subsubsection{Edge Preservation}
\begin{figure}[!t]
\centering
\includegraphics[width=3.45in]{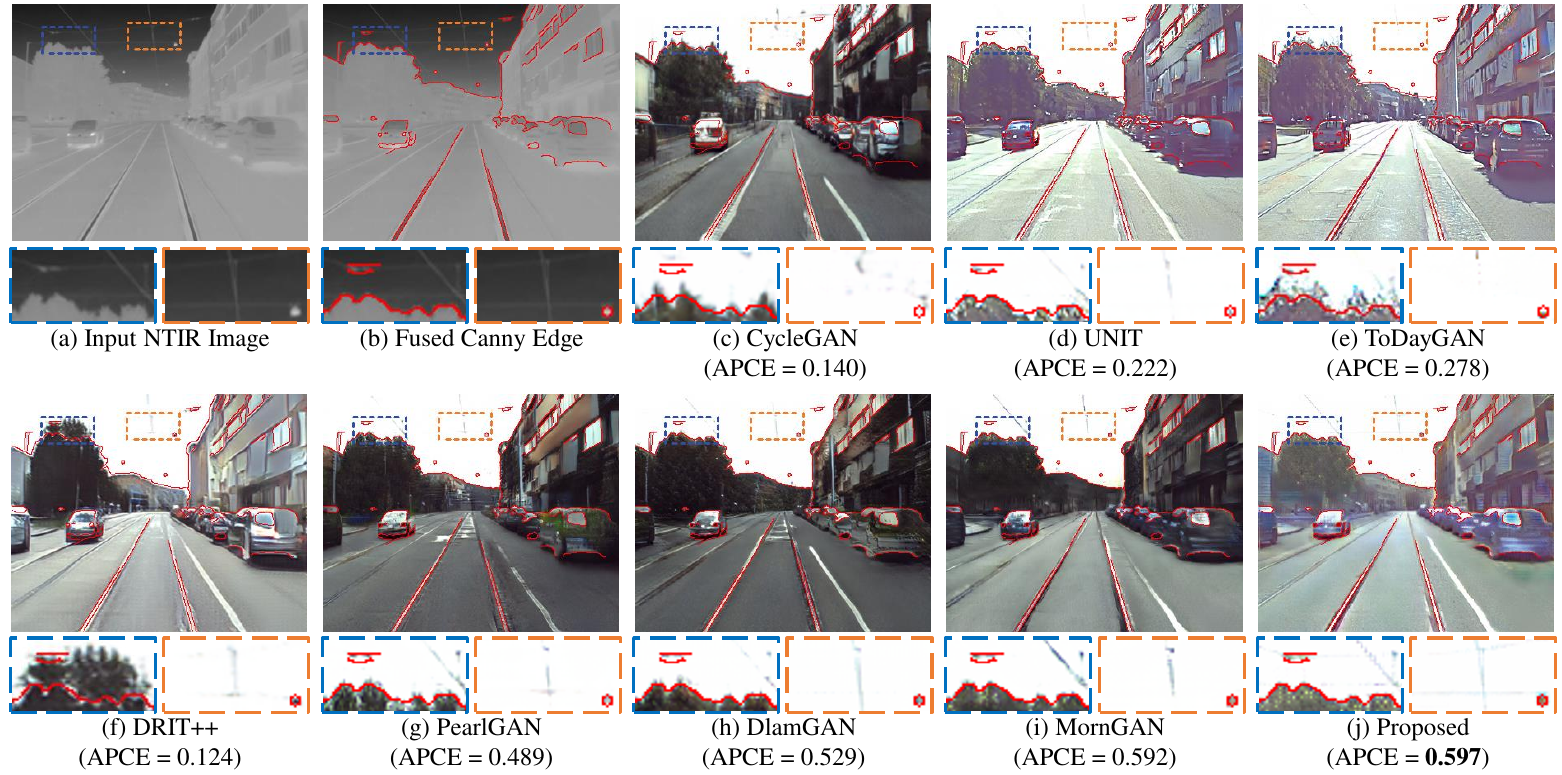}
\caption{Visual comparison of geometric consistency on the Brno dataset. The second row 
shows the enlarged results of the corresponding regions after fusion with the edges. 
The edges in red are extracted by the Canny detector from the input NTIR image.}
\label{fig_edge_brno}
\end{figure}

The qualitative comparison of edge preservation on the Brno dataset is shown in 
Fig. \ref{fig_edge_brno}. As 
shown in the blue dashed box, compared with the edges of tree in the NTIR image, the 
results of UNIT are inwardly contracted, while the results of the other compared methods 
are outwardly expanded. Unlike previous methods, the colorization result of FoalGAN can 
fit closely to the edges of the original image. For very thin wires, as shown in the 
orange dashed box, all methods except ours are unable to maintain the structure of the 
horizontal wire.

Further, the edge consistency comparison under the multi-threshold condition is shown in 
Fig. \ref{fig_apce} (b). Considering all the thresholds, the proposed method still exhibits high 
performance in the edge consistency of the translation.
\subsection{Ablation Study}

\begin{figure*}[!t]
\centering
\includegraphics[width=1\textwidth]{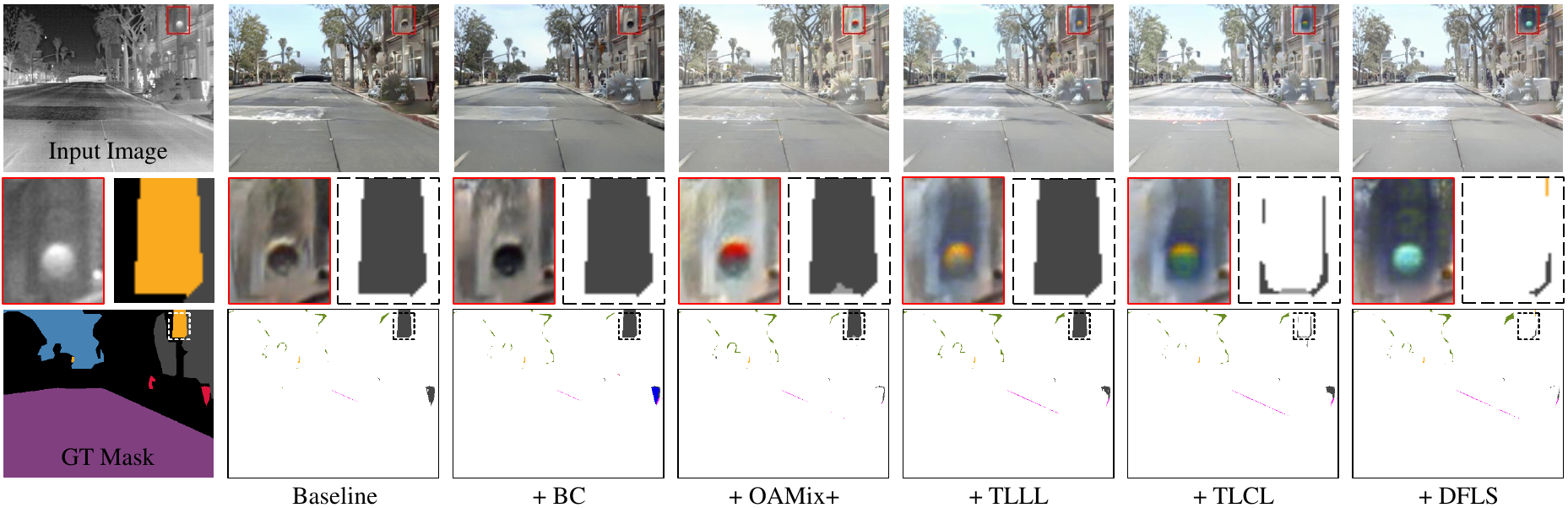}
\caption{Visual results of ablation study on the FLIR dataset. The first row shows the input NTIR 
image and the translated images by different versions of our model. In the second row, the parts covered by 
red boxes show the enlarged cropped regions in the corresponding image. The black dotted 
box is the result of zooming 
in on the corresponding area in the third row. The third row shows the error maps of the 
semantic segmentation results, where the white areas indicate the correct regions or 
unlabeled regions. The meanings of BC, OAMix+, TLLL, TLCL, and DFLS can be found in Table \ref{tab_aa}.}
\label{fig_ablation}
\end{figure*}

\begin{table}[htbp]
  \centering
  \caption{Quantitative ablation study on the FLIR dataset. ``BC" means the 
  bias correction loss and corresponding process. ``OAMix+" means the occlusion-aware mixup 
  module and corresponding appearance consistency loss. ``TLLL" means the 
  traffic light luminance loss. ``TLCL" means the traffic light color loss. ``DFLS" means 
  the dual feedback learning strategy. ``mIoU.A" and ``mIoU.S" denote the mIoU 
  results for the set of all categories and small object categories, respectively.}
  \renewcommand\tabcolsep{1.6pt}
    \begin{tabular}{ccccccccc} \toprule
    Baseline & BC    & OAMix+ & TLLL  & TLCL  & DFLS  & mIoU.A(\%) & mIoU.S(\%) & APCE \\ \hline
    \checkmark     &       &       &       &       &       & 55.7  & 21.0  & 0.58 \\
    \checkmark     & \checkmark     &       &       &       &       & 56.0  & 22.1  & 0.57 \\
    \checkmark     & \checkmark     & \checkmark     &       &       &       & 57.8  & 26.9  & 0.65 \\
    \checkmark     & \checkmark     & \checkmark     & \checkmark     &       &       & 58.1  & 27.6  & 0.65 \\
    \checkmark     & \checkmark     & \checkmark     & \checkmark     & \checkmark     &       & 58.7  & 30.0  & \textbf{0.69} \\
    \checkmark     & \checkmark     & \checkmark     & \checkmark     & \checkmark     & \checkmark     & \textbf{61.1} & \textbf{39.3} & \textbf{0.69} \\
    \bottomrule  
  \end{tabular}%
  \label{tab_aa}%
\end{table}%

Ablation analysis is performed on the FLIR dataset to discuss the validity of each 
component of FoalGAN. The results of the ablation analysis are shown in Table \ref{tab_aa}, and 
an example of a qualitative comparison is shown in Fig. \ref{fig_ablation}. We can find that the 
baseline model has poor colorization performance for traffic lights (i.e., the high temperature region 
is colored black), which misleads the segmentation model to 
identify the region as building, as shown in the second column of the figure. 
Then, due to its focus on the fake NTIR image branch, the bias correction is limited for 
the performance improvement of colorization. Moreover, as shown in Table \ref{tab_aa}, the 
edge consistency decreases slightly due to the noise in the pseudo-labels. However, 
such a bias-correction process can reduce the mismapping of the fake NTIR image branch, 
which paves the way for subsequent reduction of the context dependence of object translation.

With the introduction of the OAMix module and the appearance consistency loss, as shown in 
Table \ref{tab_aa}, the semantic consistency of SOC is significantly improved and a large gain 
in the edge consistency is achieved. And the high temperature region in the NTIR image 
is partially mapped as red, as shown in the fourth column of Fig. \ref{fig_ablation}. However, the 
lamp housing of the traffic light remains gray, which deviates from the real color distribution.

Further, as shown in the fifth column of Fig. \ref{fig_ablation}, the traffic light luminance loss can 
make the light housing color more plausible and the luminance distribution more reasonable. 
However, the color of the light in that position does not match the actual design. In the 
next experiment, traffic light color loss can reduce this color bias, i.e., by changing 
most of the yellow areas to green, as shown in the sixth column of Fig. \ref{fig_ablation}. Moreover, 
the semantic preservation of SOC gets further enhancement, as shown in Table \ref{tab_aa}.

Ultimately, as shown in the last column of Fig. \ref{fig_ablation}, the proposed dual feedback 
learning strategy not only benefits the appearance learning of traffic lights, but also 
significantly reduces the content distortion during translation. In addition, we can find 
from Table \ref{tab_aa} a massive improvement in semantic consistency of SOC, which demonstrates the 
effectiveness of the proposed strategy.
\subsection{Discussion}
In this section, we analyze $\left( 1 \right)$ the generalization capability of the proposed 
method, $\left( 2 \right)$ the color accuracy of translated traffic lights, 
and $\left( 3 \right)$ the limitations of FoalGAN.
\subsubsection{Generalization Experiments}

\begin{table}[htbp]
  \centering
  \caption{Results of generalization experiments. ``AVE.'' means average value.}
  \renewcommand\tabcolsep{2.5pt}
    \begin{tabular}{|c|c|c|c|c|c|c|}
      \hline
      & \multicolumn{3}{c|}{mIoU(\%)} & \multicolumn{3}{c|}{APCE} \\
      \cline{2-7}
          & B$\rightarrow$F   & F$\rightarrow$B   & AVE.  & B$\rightarrow$F   & F$\rightarrow$B   & AVE. \\ \hline
          \hline
    CycleGAN \cite{2017-CVPR-Zhu} & 14.3  & 19.3  & 16.8  & 0.06  & 0.07  & 0.07  \\
    UNIT \cite{2017-NIPS-Liu}  & 34.5  & 20.8  & 27.7  & 0.25  & 0.16  & 0.21  \\
    ToDayGAN \cite{2019-ICRA-Anoosheh} & 28.5  & 21.3  & 24.9  & 0.17  & 0.15  & 0.16  \\
    DRIT++ \cite{2020-IJCV-Lee} & 25.9  & 28.3  & 27.1  & 0.15  & 0.20  & 0.18  \\
    PearlGAN \cite{2022-TITS-Luo} & 45.0  & 40.7  & 42.9  & 0.42  & 0.48  & 0.45  \\
    DlamGAN \cite{2021-ICIG-Luo} & 41.1  & 43.6  & 42.4  & 0.48  & 0.54  & 0.51  \\
    MornGAN \cite{2022-Arxiv-Luo} & 47.1  & 56.0  & 51.6  & 0.43  & \textbf{0.63} & 0.53  \\
    Proposed & \textbf{51.9} & \textbf{56.1} & \textbf{54.0} & \textbf{0.57} & \textbf{0.63} & \textbf{0.60} \\
    \hline  
  \end{tabular}%
  \label{tab_genexp}%
\end{table}%

In order to explore the generalization ability of various I2I translation methods to 
out-of-domain distributions, we apply each model trained on the Brno dataset to the FLIR 
dataset abbreviated as B$\rightarrow$F and vice versa as F$\rightarrow$B. The results are 
shown in Table \ref{tab_genexp}. For the performance of semantic preservation, we can find that the 
proposed method obtains the best mIoU among all 
methods for both B$\rightarrow$F and F$\rightarrow$B experiments. Similarly, for the 
comparison of edge consistency during translation, FoalGAN consistently keeps high 
performance for APCE in both experimental modes. In summary, the results demonstrate 
the stronger generalization capability of the proposed method for domain shift.
\subsubsection{Color Accuracy of Translated Traffic Lights}

\begin{table}[htbp]
  \centering
  \caption{Comparison of the number of instances with correct color in the translated 
  traffic lights.}
    \begin{tabular}{cccc} \toprule
          & All Compared Methods & Proposed & Total Instances \\ \hline
    FLIR Dataset  & 0     & \textbf{60} & 457 \\
    Brno Dataset  & 0     & \textbf{17} & 144 \\
    \bottomrule  
  \end{tabular}%
  \label{tab_light_acc}%
\end{table}%

\begin{figure}[!t]
\centering
\includegraphics[width=3.45in]{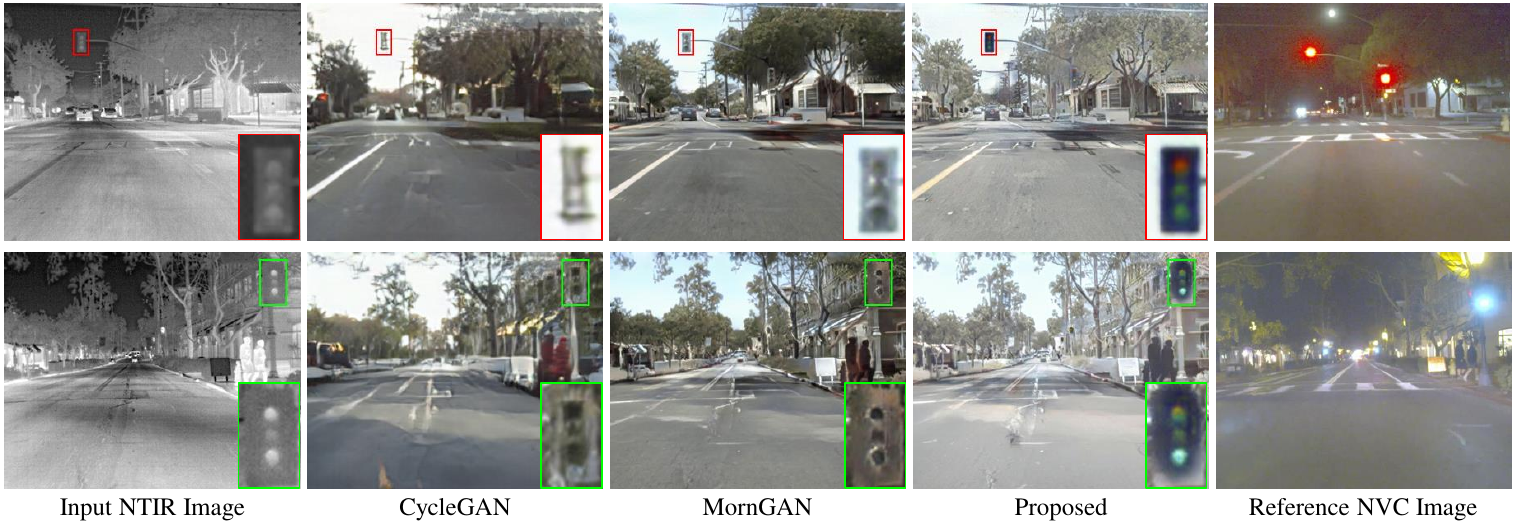}
\caption{Visual comparison of the colors of traffic light instances in the translation results.}
\label{fig_light}
\end{figure}

Since the color of signal is important for the decision making of driving systems, we 
count the color accuracy of traffic lights in the translation results. The results are 
shown in Table \ref{tab_light_acc}, where all compared I2I translation methods consistently fail to 
generate any traffic light instances with correct color, while FoalGAN can achieve 
some color accuracy on both datasets. Due to the tiny temperature difference between 
red and green lights in most NTIR images, as shown in Fig. \ref{fig_light}, it is extremely 
challenging to achieve correct traffic light colorization in the NTIR2DC task. 
Compared with the poor performance of other methods for traffic light colorization, 
the proposed method can generate plausible traffic lights and correctly infer the colors of some signals.
\subsubsection{Limitations}
\begin{figure}[!t]
\centering
\includegraphics[width=2.8in]{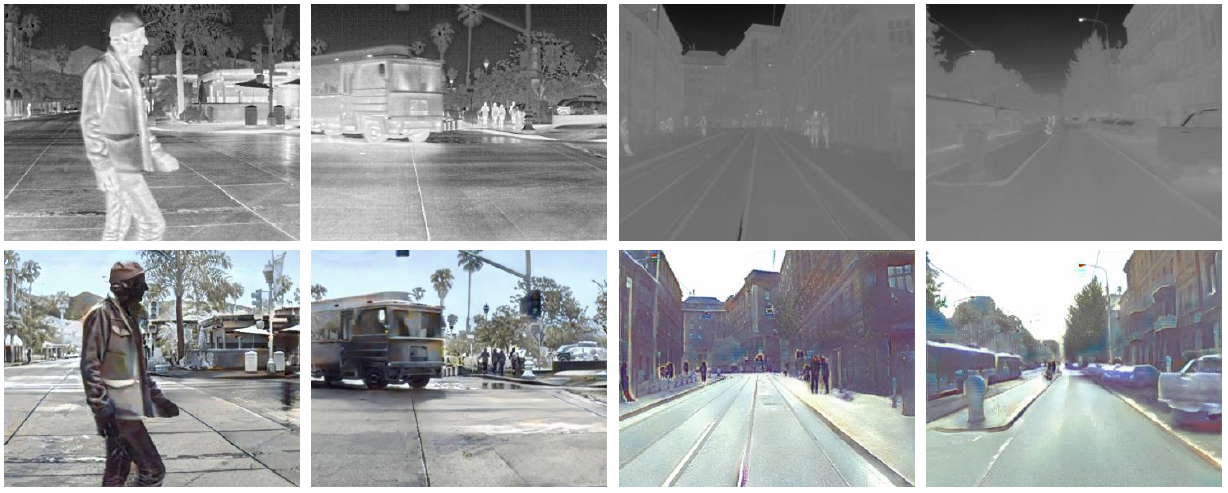}
\caption{Visualization of four failure cases. The first and second rows show the NTIR images and 
their translation results, respectively.}
\label{fig_failure}
\end{figure}

Fig. \ref{fig_failure} shows four failure cases of FoalGAN, where the input images in the first two columns 
are from the FLIR dataset, and the remaining images are from the Brno dataset. As shown in 
the first and third columns, the pedestrian's faces are unrealistic due to the diversity of 
perspectives and scales. A potential solution to this problem is to design new loss 
functions to encourage the alignment of facial features from different perspectives 
with the real features. In addition, the colorization for large objects is poor due 
to the lack of modeling for long-range dependencies, as shown in the second and fourth 
columns. To solve this problem, more attempts should be made to design corresponding 
modules to recognize large objects, and to introduce new losses to encourage the 
continuity of feature distribution in space.

% See \cite{ref1,ref2,ref3,ref4,ref5} for resources on formatting math into text and additional help in working with \LaTeX .

\section{Conclusion}
In this paper, we developed a new learning framework called FoalGAN to achieve 
colorization of NTIR images. Benefiting from the proposed dual feedback learning
strategy, the framework enabled the great improvement of the translation performance 
of small object categories. An OAMix module and the corresponding appearance 
consistency loss were proposed to reduce the context dependence of object translation. 
In addition, a traffic light appearance loss was designed to enhance the realism of 
traffic light. Moreover, we annotated a subset of the Brno dataset with pixel-wise 
category labels to further catalyze research on colorization and semantic segmentation 
of NTIR images. Comprehensive experiments demonstrate the superiority of FoalGAN for 
semantic preservation and edge consistency in the NTIR2DC task. In the future, 
designing a more generalized NTIR2DC model that can be applied to a wide range of weather 
and various imaging devices is a promising research direction.

\section*{Acknowledgments}
This work was supported by Aviation Equipment Bureau of Naval Equipment Department (21AZ0501), 
Sichuan Science and Technology Program (2022ZYD0112) and National Natural Science Foundation of China (62076055).

% \section{References Section}
% You can use a bibliography generated by BibTeX as a .bbl file.
%  BibTeX documentation can be easily obtained at:
%  http://mirror.ctan.org/biblio/bibtex/contrib/doc/
%  The IEEEtran BibTeX style support page is:
%  http://www.michaelshell.org/tex/ieeetran/bibtex/
 
%  % argument is your BibTeX string definitions and bibliography database(s)
% %\bibliography{IEEEabrv,../bib/paper}
% You can manually copy in the resultant .bbl file and set second argument of $\backslash${\tt{begin}} to the number of references
%  (used to reserve space for the reference number labels box).

% \bibliographystyle{IEEEtran}

% \end{thebibliography}
\bibliographystyle{IEEEtran}

\bibliography{refabrv}

\begin{thebibliography}{10}
\providecommand{\url}[1]{#1}
\csname url@samestyle\endcsname
\providecommand{\newblock}{\relax}
\providecommand{\bibinfo}[2]{#2}
\providecommand{\BIBentrySTDinterwordspacing}{\spaceskip=0pt\relax}
\providecommand{\BIBentryALTinterwordstretchfactor}{4}
\providecommand{\BIBentryALTinterwordspacing}{\spaceskip=\fontdimen2\font plus
\BIBentryALTinterwordstretchfactor\fontdimen3\font minus
  \fontdimen4\font\relax}
\providecommand{\BIBforeignlanguage}[2]{{%
\expandafter\ifx\csname l@#1\endcsname\relax
\typeout{** WARNING: IEEEtran.bst: No hyphenation pattern has been}%
\typeout{** loaded for the language `#1'. Using the pattern for}%
\typeout{** the default language instead.}%
\else
\language=\csname l@#1\endcsname
\fi
#2}}
\providecommand{\BIBdecl}{\relax}
\BIBdecl

\bibitem{2018-TCSVT-Li}
C.~Li, C.~Zhu, J.~Zhang, B.~Luo, X.~Wu, and J.~Tang, ``Learning local-global
  multi-graph descriptors for rgb-t object tracking,'' \emph{IEEE Trans.
  Circuits Syst. Video Technol.}, vol.~29, no.~10, pp. 2913--2926, 2018.

\bibitem{2016-TCSVT-Li}
C.~Li, X.~Wang, L.~Zhang, J.~Tang, H.~Wu, and L.~Lin, ``Weighted low-rank
  decomposition for robust grayscale-thermal foreground detection,'' \emph{IEEE
  Trans. Circuits Syst. Video Technol.}, vol.~27, no.~4, pp. 725--738, 2016.

\bibitem{2009-EOJ-W}
G.~W~Stuart and P.~K~Hughes, ``Towards an understanding of the effect of night
  vision display imagery on scene recognition,'' \emph{The Ergonomics Open
  Journal}, vol.~2, no.~1, 2009.

\bibitem{2022-TITS-Luo}
F.~Luo, Y.~Li, G.~Zeng, P.~Peng, G.~Wang, and Y.~Li, ``Thermal infrared image
  colorization for nighttime driving scenes with top-down guided attention,''
  \emph{IEEE Trans. Intell. Transp. Syst.}, vol.~23, no.~9, pp.
  15\,808--15\,823, 2022.

\bibitem{2021-ICIG-Luo}
F.~Luo, Y.~Cao, and Y.~Li, ``Nighttime thermal infrared image colorization with
  dynamic label mining,'' in \emph{International Conference on Image and
  Graphics}.\hskip 1em plus 0.5em minus 0.4em\relax Springer, 2021, pp.
  388--399.

\bibitem{2017-CVPR-Isola}
P.~Isola, J.-Y. Zhu, T.~Zhou, and A.~A. Efros, ``Image-to-image translation
  with conditional adversarial networks,'' in \emph{Proc. CVPR}, 2017, pp.
  1125--1134.

\bibitem{2014-NIPS-Goodfellow}
I.~J. Goodfellow, J.~Pouget-Abadie, M.~Mirza, B.~Xu, D.~Warde-Farley, S.~Ozair,
  A.~C. Courville, and Y.~Bengio, ``Generative adversarial nets,'' in
  \emph{Proc. NeurIPS}, 2014.

\bibitem{2017-CVPR-Zhu}
J.-Y. Zhu, T.~Park, P.~Isola, and A.~A. Efros, ``Unpaired image-to-image
  translation using cycle-consistent adversarial networks,'' in \emph{Proc.
  ICCV}, 2017, pp. 2223--2232.

\bibitem{2018-ECCV-Huang}
X.~Huang, M.-Y. Liu, S.~Belongie, and J.~Kautz, ``Multimodal unsupervised
  image-to-image translation,'' in \emph{Proc. ECCV}, 2018, pp. 172--189.

\bibitem{2020-IJCV-Lee}
H.-Y. Lee, H.-Y. Tseng, Q.~Mao, J.-B. Huang, Y.-D. Lu, M.~Singh, and M.-H.
  Yang, ``Drit++: Diverse image-to-image translation via disentangled
  representations,'' \emph{International Journal of Computer Vision}, vol. 128,
  no.~10, pp. 2402--2417, 2020.

\bibitem{2019-ICRA-Anoosheh}
A.~Anoosheh, T.~Sattler, R.~Timofte, M.~Pollefeys, and L.~Van~Gool,
  ``Night-to-day image translation for retrieval-based localization,'' in
  \emph{ICRA}, 2019, pp. 5958--5964.

\bibitem{2018-ICLR-Zhang}
H.~Zhang, M.~Cisse, Y.~N. Dauphin, and D.~Lopez-Paz, ``mixup: Beyond empirical
  risk minimization,'' in \emph{Proc. ICLR}, 2018.

\bibitem{2005-LI-Vollmeyer}
R.~Vollmeyer and F.~Rheinberg, ``A surprising effect of feedback on learning,''
  \emph{Learning and instruction}, vol.~15, no.~6, pp. 589--602, 2005.

\bibitem{2014-BBR-Luft}
C.~D.~B. Luft, ``Learning from feedback: The neural mechanisms of feedback
  processing facilitating better performance,'' \emph{Behavioural brain
  research}, vol. 261, pp. 356--368, 2014.

\bibitem{2020-ICRA-Ligocki}
A.~Ligocki, A.~Jelinek, and L.~Zalud, ``Brno urban dataset-the new data for
  self-driving agents and mapping tasks,'' in \emph{Proc. ICRA}.\hskip 1em plus
  0.5em minus 0.4em\relax IEEE, 2020, pp. 3284--3290.

\bibitem{2018-Berg-CVPRW}
A.~Berg, J.~Ahlberg, and M.~Felsberg, ``Generating visible spectrum images from
  thermal infrared,'' in \emph{Proc. CVPR Workshops}, 2018, pp. 1143--1152.

\bibitem{2020-Bhat-ICCES}
N.~Bhat, N.~Saggu, S.~Kumar \emph{et~al.}, ``Generating visible spectrum images
  from thermal infrared using conditional generative adversarial networks,'' in
  \emph{ICCES}, 2020, pp. 1390--1394.

\bibitem{2020-Kuang-IPT}
X.~Kuang, J.~Zhu, X.~Sui, Y.~Liu, C.~Liu, Q.~Chen, and G.~Gu, ``Thermal
  infrared colorization via conditional generative adversarial network,''
  \emph{Infrared Physics \& Technology}, p. 103338, 2020.

\bibitem{2021-ISEE-Le}
T.~Le-Tien, T.~H.~D. Quang, H.~Y. Vy, T.~Nguyen-Thanh, and H.~Phan-Xuan,
  ``Gan-based thermal infrared image colorization for enhancing object
  identification,'' in \emph{Proc. ISEE}.\hskip 1em plus 0.5em minus
  0.4em\relax IEEE, 2021, pp. 90--94.

\bibitem{2020-TCSVT-Zhao}
Y.~Zhao, L.-M. Po, K.-W. Cheung, W.-Y. Yu, and Y.~A.~U. Rehman, ``Scgan:
  saliency map-guided colorization with generative adversarial network,''
  \emph{IEEE Trans. Circuits Syst. Video Technol.}, vol.~31, no.~8, pp.
  3062--3077, 2020.

\bibitem{2018-Nyberg-ECCV}
A.~Nyberg, A.~Eldesokey, D.~Bergstrom, and D.~Gustafsson, ``Unpaired thermal to
  visible spectrum transfer using adversarial training,'' in \emph{Proc. ECCV},
  2018, pp. 0--0.

\bibitem{2022-Arxiv-Luo}
F.-Y. Luo, Y.-J. Cao, K.-F. Yang, and Y.-J. Li, ``Memory-guided collaborative
  attention for nighttime thermal infrared image colorization,'' \emph{arXiv
  preprint arXiv:2208.02960}, 2022.

\bibitem{2017-NIPS-Liu}
M.-Y. Liu, T.~Breuel, and J.~Kautz, ``Unsupervised image-to-image translation
  networks,'' in \emph{Proc. NeurIPS}, 2017.

\bibitem{2020-ECCV-Zheng}
Z.~Zheng, Y.~Wu, X.~Han, and J.~Shi, ``Forkgan: Seeing into the rainy night,''
  in \emph{Proc. ECCV}.\hskip 1em plus 0.5em minus 0.4em\relax Springer, 2020,
  pp. 155--170.

\bibitem{2020-CVPR-Bhattacharjee}
D.~Bhattacharjee, S.~Kim, G.~Vizier, and M.~Salzmann, ``Dunit: Detection-based
  unsupervised image-to-image translation,'' in \emph{Proc. CVPR}, 2020, pp.
  4787--4796.

\bibitem{2021-TCSVT-Fu}
L.~Fu, H.~Yu, F.~Juefei-Xu, J.~Li, Q.~Guo, and S.~Wang, ``Let there be light:
  Improved traffic surveillance via detail preserving night-to-day transfer,''
  \emph{IEEE Trans. Circuits Syst. Video Technol.}, 2021.

\bibitem{2018-ECCV-Huang-Auggan}
S.-W. Huang, C.-T. Lin, S.-P. Chen, Y.-Y. Wu, P.-H. Hsu, and S.-H. Lai,
  ``Auggan: Cross domain adaptation with gan-based data augmentation,'' in
  \emph{Proc. ECCV}, 2018, pp. 718--731.

\bibitem{2019-WACV-Cherian}
A.~Cherian and A.~Sullivan, ``Sem-gan: semantically-consistent image-to-image
  translation,'' in \emph{Proc. WACV}.\hskip 1em plus 0.5em minus 0.4em\relax
  IEEE, 2019, pp. 1797--1806.

\bibitem{2020-TCSVT-Tan}
D.~S. Tan, Y.-X. Lin, and K.-L. Hua, ``Incremental learning of multi-domain
  image-to-image translations,'' \emph{IEEE Trans. Circuits Syst. Video
  Technol.}, vol.~31, no.~4, pp. 1526--1539, 2020.

\bibitem{1999-IP-Ford}
A.~Ford and F.~A. Ford, \emph{Modeling the environment: an introduction to
  system dynamics models of environmental systems}.\hskip 1em plus 0.5em minus
  0.4em\relax Island press, 1999.

\bibitem{1998-Nature-Hupe}
J.~Hup{\'e}, A.~James, B.~Payne, S.~Lomber, P.~Girard, and J.~Bullier,
  ``Cortical feedback improves discrimination between figure and background by
  v1, v2 and v3 neurons,'' \emph{Nature}, vol. 394, no. 6695, pp. 784--787,
  1998.

\bibitem{2007-Neuron-Gilbert}
C.~D. Gilbert and M.~Sigman, ``Brain states: top-down influences in sensory
  processing,'' \emph{Neuron}, vol.~54, no.~5, pp. 677--696, 2007.

\bibitem{2012-JCogNeu-Wyatte}
D.~Wyatte, T.~Curran, and R.~O'Reilly, ``The limits of feedforward vision:
  Recurrent processing promotes robust object recognition when objects are
  degraded,'' \emph{Journal of Cognitive Neuroscience}, vol.~24, no.~11, pp.
  2248--2261, 2012.

\bibitem{2014-NeurIPS-Stollenga}
M.~F. Stollenga, J.~Masci, F.~Gomez, and J.~Schmidhuber, ``Deep networks with
  internal selective attention through feedback connections,'' \emph{Proc.
  NeurIPS}, vol.~27, 2014.

\bibitem{2017-CVPR-Zamir}
A.~R. Zamir, T.-L. Wu, L.~Sun, W.~B. Shen, B.~E. Shi, J.~Malik, and
  S.~Savarese, ``Feedback networks,'' in \emph{Proc. CVPR}, 2017, pp.
  1308--1317.

\bibitem{2019-CVPR-Li}
Z.~Li, J.~Yang, Z.~Liu, X.~Yang, G.~Jeon, and W.~Wu, ``Feedback network for
  image super-resolution,'' in \emph{Proc. CVPR}, 2019, pp. 3867--3876.

\bibitem{2022-TCSVT-Song}
X.~Song, D.~Zhou, W.~Li, H.~Ding, Y.~Dai, and L.~Zhang, ``Wsamf-net: Wavelet
  spatial attention based multi-stream feedback network for single image
  dehazing,'' \emph{IEEE Trans. Circuits Syst. Video Technol.}, 2022.

\bibitem{2019-ICCV-Yun}
S.~Yun, D.~Han, S.~J. Oh, S.~Chun, J.~Choe, and Y.~Yoo, ``Cutmix:
  Regularization strategy to train strong classifiers with localizable
  features,'' in \emph{Proc. ICCV}, 2019, pp. 6023--6032.

\bibitem{2021-WACV-Olsson}
V.~Olsson, W.~Tranheden, J.~Pinto, and L.~Svensson, ``Classmix:
  Segmentation-based data augmentation for semi-supervised learning,'' in
  \emph{Proc. WACV}, 2021, pp. 1369--1378.

\bibitem{2022-TCSVT-Zhou}
Q.~Zhou, Z.~Feng, Q.~Gu, J.~Pang, G.~Cheng, X.~Lu, J.~Shi, and L.~Ma,
  ``Context-aware mixup for domain adaptive semantic segmentation,'' \emph{IEEE
  Trans. Circuits Syst. Video Technol.}, 2022.

\bibitem{2019-TCSVT-Takahashi}
R.~Takahashi, T.~Matsubara, and K.~Uehara, ``Data augmentation using random
  image cropping and patching for deep cnns,'' \emph{IEEE Trans. Circuits Syst.
  Video Technol.}, vol.~30, no.~9, pp. 2917--2931, 2019.

\bibitem{2019-detectron2-Wu}
Y.~Wu, A.~Kirillov, F.~Massa, W.-Y. Lo, and R.~Girshick, ``Detectron2,''
  \url{https://github.com/facebookresearch/detectron2}, 2019.

\bibitem{2020-Arxiv-Tao}
A.~Tao, K.~Sapra, and B.~Catanzaro, ``Hierarchical multi-scale attention for
  semantic segmentation,'' \emph{arXiv preprint arXiv:2005.10821}, 2020.

\bibitem{2017-ICCV-Neuhold}
G.~Neuhold, T.~Ollmann, S.~Rota~Bulo, and P.~Kontschieder, ``The mapillary
  vistas dataset for semantic understanding of street scenes,'' in \emph{Proc.
  ICCV}, 2017, pp. 4990--4999.

\bibitem{2004-TIP-Wang}
Z.~Wang, A.~C. Bovik, H.~R. Sheikh, and E.~P. Simoncelli, ``Image quality
  assessment: from error visibility to structural similarity,'' \emph{IEEE
  Trans. Image Process.}, vol.~13, no.~4, pp. 600--612, 2004.

\bibitem{2015-CVPR-Hwang}
S.~Hwang, J.~Park, N.~Kim, Y.~Choi, and I.~So~Kweon, ``Multispectral pedestrian
  detection: Benchmark dataset and baseline,'' in \emph{Proc. CVPR}, 2015, pp.
  1037--1045.

\bibitem{2019-FLIR-FA}
F.A.Group, ``Flir thermal dataset for algorithm training,''
  \url{https://www.flir.co.uk/oem/adas/adas-dataset-form/}, May 2019.

\bibitem{2015-IJCV-Everingham}
M.~Everingham, S.~A. Eslami, L.~Van~Gool, C.~K. Williams, J.~Winn, and
  A.~Zisserman, ``The pascal visual object classes challenge: A
  retrospective,'' \emph{International journal of computer vision}, vol. 111,
  no.~1, pp. 98--136, 2015.

\bibitem{2010-IJCV-Everingham}
M.~Everingham, L.~Van~Gool, C.~K. Williams, J.~Winn, and A.~Zisserman, ``The
  pascal visual object classes (voc) challenge,'' \emph{International journal
  of computer vision}, vol.~88, no.~2, pp. 303--338, 2010.

\bibitem{2014-Arxiv-Kingma}
D.~P. Kingma and J.~Ba, ``Adam: A method for stochastic optimization,'' in
  \emph{ICLR (Poster)}, 2015.

\bibitem{2016-CVPR-Cordts}
M.~Cordts, M.~Omran, S.~Ramos, T.~Rehfeld, M.~Enzweiler, R.~Benenson,
  U.~Franke, S.~Roth, and B.~Schiele, ``The cityscapes dataset for semantic
  urban scene understanding,'' in \emph{Proc. CVPR}, 2016, pp. 3213--3223.

\bibitem{2022-Arxiv-Wang}
C.-Y. Wang, A.~Bochkovskiy, and H.-Y.~M. Liao, ``Yolov7: Trainable
  bag-of-freebies sets new state-of-the-art for real-time object detectors,''
  \emph{arXiv preprint arXiv:2207.02696}, 2022.

\bibitem{2014-ECCV-Lin}
T.-Y. Lin, M.~Maire, S.~Belongie, J.~Hays, P.~Perona, D.~Ramanan,
  P.~Doll{\'a}r, and C.~L. Zitnick, ``Microsoft coco: Common objects in
  context,'' in \emph{Proc. ECCV}, 2014, pp. 740--755.

\end{thebibliography}

\newpage

 

% \end{IEEEbiography}

\vspace{11pt}


\vfill

\end{document}